\documentclass[10pt,journal,twocolumn,twoside]{IEEEtran} 

\usepackage{graphicx}
\usepackage{subcaption}
\usepackage{epstopdf}
\usepackage{float}
\usepackage{algorithmic}
\usepackage{array}
\usepackage{amsmath}
\usepackage{amssymb}
\usepackage{mdwmath}
\usepackage{mdwtab}
\usepackage{eqparbox}
\usepackage{stfloats}
\usepackage{fixltx2e}
\usepackage{hyperref}
\usepackage{cases} 

\usepackage{flushend}
\usepackage{upgreek}

\usepackage{booktabs}
\usepackage{multirow}

\usepackage{xcolor}
\usepackage{makecell}
\usepackage[boxed,ruled,commentsnumbered]{algorithm2e}
\usepackage{url}
\usepackage{cite}

\usepackage{textcomp}
\usepackage{bm}
\usepackage{setspace}
\usepackage{epsfig}
\usepackage{diagbox}

\allowdisplaybreaks[4]

\makeatletter

\renewcommand*{\@opargbegintheorem}[3]{\trivlist
      \item[\hskip \labelsep{\bfseries #1\ #2}] \textbf{(#3):}\ }
\makeatother     

\begin{document}

\title
{Task-Oriented Communication for Graph Data: A Graph Information Bottleneck Approach}
\author{ Shujing Li, Yanhu Wang, Shuaishuai Guo, ~\IEEEmembership{Senior Member, IEEE}, and Chenyuan Feng, ~\IEEEmembership{Member, IEEE}
\thanks{ Shujing Li, Yanhu Wang and Shuaishuai Guo are with School of Control Science and Engineering, Shandong University, Jinan 250061, China and also with Shandong Key Laboratory of Wireless Communication Technologies, Shandong University, Jinan, Shandong, P. R. China (e-mail: lishujing777@mail.sdu.edu.cn, yh-wang@mail.sdu.edu.cn, shuaishuai\_guo@mail.sdu.edu.cn).}
\thanks{Chenyuan Feng is with Department of Communication Systems, EURECOM, Sophia Antipolis 06410, France (e-mail: Chenyuan.Feng@eurecom.fr).}}
\maketitle

\begin{abstract} 

Graph data, essential in fields like knowledge representation and social networks, often involves large networks with many nodes and edges. Transmitting these graphs can be highly inefficient due to their size and redundancy for specific tasks. This paper introduces a method to extract a smaller, task-focused subgraph that maintains key information while reducing communication overhead. Our approach utilizes graph neural networks (GNNs) and the graph information bottleneck (GIB) principle to create a compact, informative, and robust graph representation suitable for transmission. The challenge lies in the irregular structure of graph data, making GIB optimization complex. We address this by deriving a tractable variational upper bound for the objective function. Additionally, we propose the VQ-GIB mechanism, integrating vector quantization (VQ) to convert subgraph representations into a discrete codebook sequence, compatible with existing digital communication systems. Our experiments show that this GIB-based method significantly lowers communication costs while preserving essential task-related information. The approach demonstrates robust performance across various communication channels, suitable for both continuous and discrete systems.

\end{abstract}


\section{Introduction} 
\IEEEPARstart{I}{n} the swiftly changing realm of communication systems, marked by the emergence of 5G and its advancements, an increasing demand arises for efficient and smart communication frameworks capable of adjusting to intricate and ever-shifting environments. Conventional communication systems concentrate on enhancing metrics such as throughput and latency, frequently neglecting the distinct needs of task-oriented communications. In task-oriented scenarios, the objective extends beyond the mere effective transmission of raw data; it aims to optimize the performance for specific tasks like classification, prediction, or decision-making. It is worth noting that in addition to Euclidean data such as images and text, there are vast amounts of non-Euclidean data generated in social networks, biological networks, transportation networks and recommendation systems, etc. \cite{ref002} This graph-structured data has no order or coordinate reference points, and is difficult to represent in a grid-like matrix or a tensor. A graph is usually composed of nodes and edges, which contain rich relational information. We can regard the graph data as a web of relationships, wherein the nodes are the subjects and the edges represent the relationships between the nodes. This feature makes graph an effective tool for modeling complex relationships in various non-Euclidean data scenarios \cite{9027556,7976923,8047276,7965747,4684911}. 

Graph data is widely applied to knowledge representation, recommendation systems, and user behavior analysis \cite{ref001}. However, the inherent complexity and size of graph data can pose challenges for transmission and storage. A complete graph, which contains all possible relationships between the nodes, is often not suitable for direct transmission due to the substantial bandwidth and storage resources required. Moreover, a complete graph tends to contain redundant information, leading to unnecessary resource consumption. In many practical applications, only specific task-related information within the graph needs to be transmitted to accomplish particular objectives. Additionally, graph data might contain sensitive information or personal privacy that should be safeguarded during transmission. Transmitting the entire graph increases the risk of information leakage. Therefore, an intelligent and concise communication system is needed to transmit graph data effectively. Within this context, the Graph Information Bottleneck (GIB) stands out as an innovative strategy to tackle these challenges. It draws upon the principles of information theory and graph neural networks to craft solutions. Central to this approach is the compression of transmitted information, ensuring that it retains only the most pertinent features relevant to the task. This targeted compression facilitates a more efficient and impactful mode of communication.

\subsection{Related Works}
Task-oriented communication is a promising solution for graph data transmission. Different from traditional communication, it focuses on the accurate transmission of task-relevant information, rather than the bit-level precise transmission  \cite{Gndz2022BeyondTB, Xie2020DeepLE}. Its characteristics align well with our demands for graph data transmission. It is difficult to characterize or extract task-related information with mathematical models, and the existing task-oriented communication systems mainly rely on deep learning technology \cite{Luo2022SemanticCO}. That is, the neural networks (NNs)-based encoder and decoder are constructed and trained, so that the task-oriented communication system obtains the ability to extract task-relevant information \cite{Shao2021LearningTC, Weng2021SemanticCF, Shao2021TaskOrientedCF}.

Precisely, Farsad et al. proposed a neural network architecture for text transmission task, that combines joint source-channel coding (JSCC) with recurrent neural network (RNN) based encoder, binarization layers, channel layers, and RNN-based decoder \cite{Farsad2018DeepLF}. Xie et al. introduced a system for textual semantic restoration tasks based on Transformer model \cite{Xie2020DeepLE}. Their approach focuses on maximizing system capacity while minimizing semantic errors by restoring the meaning of sentences. Guo et al. proposed to utilize pre-trained language models to quantify the semantic importance of text and allocate unequal power based on semantic importance \cite{10177738}. Some JSCC methods map image pixel values directly to input representations to achieve high-quality image reconstruction tasks \cite{Bourtsoulatze2018DeepJS,Kurka2019DeepJSCCfDJ}. Kang et al. proposed an image transmission method specifically designed for scene classification tasks \cite{Kang2021TaskOrientedIT}. Shao et al. developed a task-oriented communication scheme for edge inference using the information bottleneck (IB) \cite{Shao2021LearningTC,Shao2021TaskOrientedCF}. Their scheme aims to improve the performance of image classification task at the edge server by efficiently transmitting relevant information. These NNs-based task-oriented communication systems have achieved outstanding outcomes, demonstrating  the capacity to effectively execute specific tasks.

\subsection{Motivation \& Contributions}
However, these works cannot be directly used for graph data transmission.
Because they are based on CNNs or fully connected networks and deal with regular data such as text (represented as sequences of characters) and images (represented as continuous two-dimensional or three-dimensional pixel sets).
Graph data exhibits a more intricate structure, which is composed of nodes and edges with attribute information.
When transmitting graph data, it is important to consider the efficiency of transmitting large-scale graph data while accurately preserving the correlation between nodes and edges \cite{Li2022GraphSC, 9416834}.
Therefore, it is necessary to design a new task-oriented communication system for graph data transmission.
Using information bottleneck theory to develop task-oriented communication systems is a good option \cite{Shao2021LearningTC, Wang2023PrivacyPreservingTS}.
Because IB theory aims to find an optimal intermediate representation that can retain important information in the input data to ensure the accuracy of the output prediction while eliminating redundant information \cite{DBLP:journals/corr/physics-0004057, Tishby2015DeepLA}. 
However, it should be noted that the direct utilization of IB-based frameworks in graph data processing is not feasible. This is because the IB framework assumes that the data follows an independent identically distributed (IID) pattern.
Whereas in graph data, the presence of edges and their attributes results in the data points being dependent on one another, which makes the graph data deviate from the IID assumption \cite{Wu2022HandlingDS,MACFL2022TWC}. 

Recently, an information-theoretical design principle for graph data, named GIB, has been developed, which seeks the right balance between data compression and information preservation for graph data\cite{DBLP:conf/nips/WuRLL20}. Specifically, employing GNNs\footnote{GNN is a deep learning model specifically designed to work with graph data.  In contrast to traditional deep learning models primarily handling vectorized data, GNNs excel at capturing intricate relationships within graphs, involving nodes and edges\cite{Wu2019ACS}. The fundamental principle of GNNs is to gradually aggregate local neighborhood information by iteratively updating the representation of nodes\cite{LargeGNN}. Typical GNN models include Graph Convolutional Networks (GCNs), Graph Isomorphism Network (GIN), Graph Attention Networks (GATs), etc.} as the foundational framework for graph data processing, GIB works well. Through the process of learning and parameter adjustments, GIB ensures the preservation of only task-relevant information while simultaneously compressing extraneous data.
One of the challenges in studying graph information bottleneck, i.e., how to handle the interdependencies between nodes in graph data, is the non-IID character of graph data. One of the primary technical contributions of our research is task-oriented transmission, which we will further optimize for.

In addition, the actual widespread use of digital communication systems makes the realization of compatibility between task-oriented communication systems and digital communication systems a necessity.
Thus, there is a requirement to identify a suitable digitization mechanism for communication systems geared towards graph data, ensuring both effective data compression and robustness. Vector quantization (VQ) emerges as a powerful technique that addresses these needs by mapping high-dimensional data into a finite set of lower-dimensional codewords. This process not only facilitates significant data compression but also enhances the system's robustness against noise and transmission errors. Furthermore, by integrating with deep learning, VQ is highly adaptive and can be dynamically adjusted to different data distributions and channel conditions.

Motivated by these issues, we design GIB-enabled task-oriented communication systems for graph data in this work. Our main contributions are summarized as follows:

\begin{itemize}
    \item We introduce GIB into a task-oriented communication system for graph data. We build a Markov chain model for information transmission and formulate an optimization problem to maximize the mutual information between the task target and received codewords while minimizing the mutual information between the received codewords and the raw graph data. This approach balances the preservation of critical information in the extracted features while eliminating redundant information, thereby enhancing task success rates and reducing communication overhead.
    \item To address the difficulty of handling mutual information terms in GIB due to high-dimensional integration, we use the Mutual Information Neural Estimator (MINE) to directly estimate the mutual information between the original graph and the subgraphs. This approach overcomes the challenge of obtaining the prior distribution of the subgraphs in applying variational approximation methods.
    \item Recognizing the importance of topological information in graph data, particularly in revealing community structures, we introduce a connectivity loss term into the objective function. This term leverages topological information during feature extraction, reduces fluctuations in ambiguous node assignments, and contributes to a more stable training process.
    \item We map the resulting subgraph representation onto a jointly trained codebook to generate a discrete index sequence for transmission.  This mapping ensures compatibility with existing digital communication systems.  Experimental results show that the system is able to achieve higher compression rates while achieving task success rates comparable to traditional digital communication methods.
\end{itemize}

\subsection{Organization}
The subsequent sections of this paper are organized as follows.
In Section \uppercase\expandafter{\romannumeral2}, we outline the system model, describe the structure and design objectives of the task-oriented communication system for graph data. Section \uppercase\expandafter{\romannumeral3} introduces the details of GIB-enabled task-oriented communication systems for graph data, including the handling of GIB objectives and the system training strategy. Section \uppercase\expandafter{\romannumeral4} presents our proposed approach for digitizing the task-oriented communication system. In Section \uppercase\expandafter{\romannumeral5}, we evaluate the performance and effectiveness of the proposed method through experiments. In Section \uppercase\expandafter{\romannumeral6}, potential applications of the proposed method in practical scenarios are discussed. Finally, we make a brief summary of this paper in Section \uppercase\expandafter{\romannumeral7}.

\subsection{Notations}
In this paper, a graph with $m$ nodes is defined as $g$ or $G=\left( V,E,A,\mathcal{X} \right)$, where $V=\left\{ {V_{i}}|i=1,2,...,m \right\}$ is the set of nodes with cardinality $m$, $E=\left\{ \left( {V_{i}},{V_{j}} \right)|i<j,\text{ }{V_{i}}\text{ }and\text{ }{V_{j}}\text{ } are\text{ }connected \right\}$ is the edge set, $A={{\left\{ 0,1 \right\}}^{m\times m}}$ is the adjacent matrix, and $\mathcal{X}\in {{R}^{m\times d}}$ is the feature matrix corresponding to $V$ with feature dimension $d$. The pair $\left( G,Y \right)$ stands for the graph data and its target variable. 
The entropy of $Y$ is defined as $H\left( Y \right)$. The mutual information between $X$ and $Y$ is represented as $I\left( X,Y \right)$. 

\begin{figure*}[htbp]
    \centering
    \includegraphics[width=\textwidth]{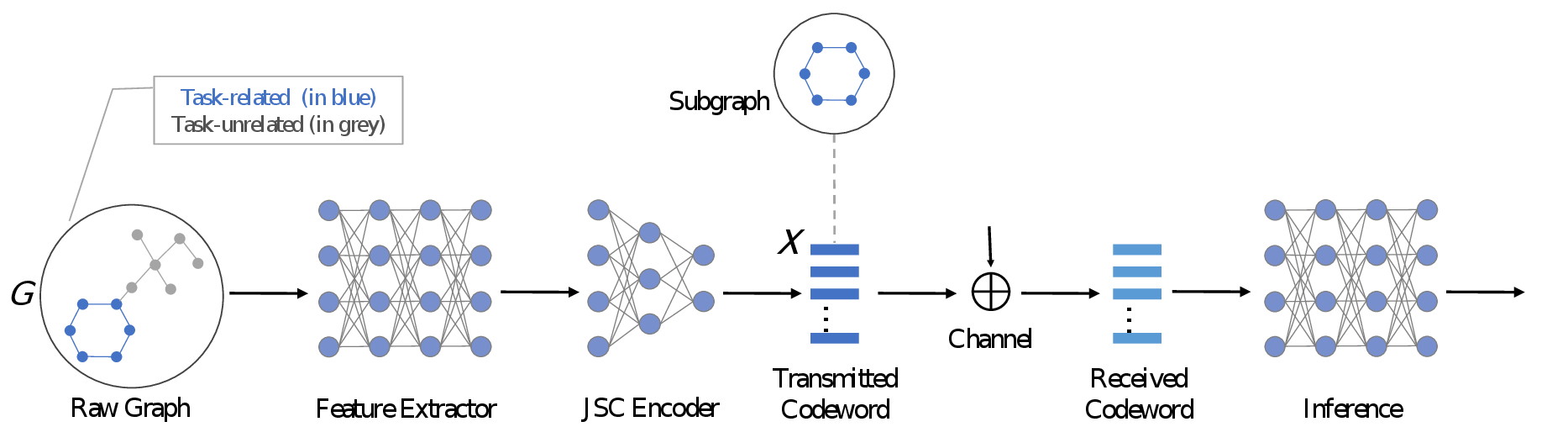}
    \caption{ Task-Oriented Communication Scheme for Graph Data: In this scheme, the \textquotedblleft transmitted codeword \textquotedblright refers to the encoded representation of the task-related subgraph, while the \textquotedblleft corrupted codeword \textquotedblright denotes the representation of the subgraph received by the receiver, which has undergone corruption during transmission through the channel. The symbol $\hat{Y}$
  represents the task inference output.}
    \label{fig1}
    \end{figure*} 

\section{System Model and Problem Description}
\subsection{System Model}
In this paper, we consider a task-oriented communication system for graph data as shown in Fig. \ref{fig1}. The system mainly includes a transmitter and a receiver, where the transmitter consists of a feature extractor implemented by a GNN and a joint source-channel (JSC) encoder. The receiver is a neural network used for task inference. Given an input graph $g$, they cooperate to perform tasks to make the inference output $\hat{y}$ consistent with the true target variable $y$.
The random variables $(G, Y)$ together with the encoded codeword $X$ and the channel-corrupted codeword $\hat X$ form the following probabilistic model:
\begin{equation}
\label{eq1}
\left( Y \right)G \to X \to \hat X \to \hat Y,
\end{equation}
which satisfy $p\left( \hat{y}|g \right)={{p}_{\theta }}\left( \hat{y}|\hat{x} \right){{p}_{channel}}\left( \hat{x}|x \right){p}_{\phi }\left( x|g \right)$.
Herein, the lowercase letters $g$, $x$, $\hat{x}$, $y$, and $\hat{y}$ denote realizations of the variables represented by the corresponding uppercase letters.
${{p}_{\phi }}\left( x|g \right)$ represents the transmitter neural network parameterized by adjustable parameters $\phi$.
For a given input graph $g$, the feature encoder identifies the representation $x$ related to the task. The task-relevant features are encoded by the JSC encoder and then transmitted to the receiver through the channel. 

To accommodate both continuous channels and discrete channels within this framework, we introduce an indicator variable $\zeta$ that distinguishes between the channel types: $\zeta=0$ for an continuous channel and $\zeta=1$ for an discrete. The conditional probability distribution ${{p}_{channel}}\left( \hat{x}|x \right)$ is thus defined as:
\begin{equation}
\label{unifiedChannel}
p_{\text{channel}}(\hat{x}|x; \zeta) =
\begin{cases}
    \frac{1}{\sqrt{2\pi N_0}}\exp\left(-\frac{(\hat{x}-x)^2}{2N_0}\right), & \text{if } \zeta = 0 \\
    P_{\hat{X}|X}(\hat{x}|x). & \text{if } \zeta = 1
\end{cases}
\end{equation}
where we assume that the continuous channel is the additive white Gaussian noise (AWGN) channel and the discrete channel is the Symmetric Discrete Channel (SDC). 

AWGN is implemented using an untrained neural network layer, and the transfer function is expressed as:

\begin{equation}
\label{eq2}
    \hat{x}=x+\epsilon,
\end{equation}
where the Gaussian noise $\epsilon \sim \mathcal{N}\left( 0,{{N_0}/2} \right)$.

For SDC, $P_{\hat{X}|X}(\hat{x}|x)$ signifies the transition probability matrix, encapsulating the likelihood of transitioning from input to output symbols.
The SDC model assumes that both channel inputs and outputs utilize the same symbol set, with each symbol corresponding to an identical output probability distribution. This probabilistic behavior is succinctly captured by a transition matrix:
\begin{equation}
\label{eq27}
    \mathcal{P}={{\left[ \begin{matrix}
   \varepsilon  & \frac{1-\varepsilon }{r-1} & \cdots  & \frac{1-\varepsilon }{r-1}  \\
   \frac{1-\varepsilon }{r-1} & \varepsilon  & \cdots  & \frac{1-\varepsilon }{r-1}  \\
   \cdots  & \cdots  & \cdots  & \cdots  \\
   \frac{1-\varepsilon }{r-1} & \frac{1-\varepsilon }{r-1} & \cdots  & \varepsilon   \\
\end{matrix} \right]}_{r\times r}},
\end{equation}
where $\varepsilon $ represents the probability of correct symbol transmission and 
$r$ denotes the cardinality of the channel symbol set. Potential transmission errors could lead to the receiver retrieving incorrect vectors, impacting the inference task.

The conditional distribution ${{p}_{\theta }}\left( \hat{y}|\hat{x} \right)$ stands for the task inference network parameterized by adjustable parameters $\theta $ at the receiver.
 It deduces the class label $\hat{y}$  based on $\hat{x}$ based on the channel-corrupted codeword $\hat{x}$.

\subsection{Problem Description}
The effectiveness of task execution is intricately linked to the dimension of the feature vector produced by the transmitter. For instance, in the context of graph classification tasks, a higher dimension of the feature vector could lead to an elevated classification accuracy. However, this advantage comes at the cost of increased communication overhead. Thus, the challenge is to identify a concise and informative representation that aligns with the optimal subgraph.
To formalize this trade-off, we formulate the following objective function based on the GIB principle:
\begin{equation}
\label{eq3}
{{\mathcal{L}}_{GIB}}=-I\left( Y,\hat{X} \right)+\beta I\left( G,\hat{X} \right),
\end{equation}
in which $I(Y,\hat{X})$ represents the mutual information capturing the relevance of task-specific information in the received codeword, $I(G,\hat{X})$ represents the preserved information in $\hat{X}$ given $G$, and $\beta$ acts as a trade-off factor governing the relationship between the two.

Developing task-oriented communication based on GIB is a promising approach to solving the graph transmission challenges. 
This approach offers an information metric for the graph data and can effectively capture node and edge information.  However, there are a few noteworthy problems:

\begin{itemize}
\item \textbf{Problem 1}: How to deal with mutual information containing graph data to get a tractable objective function?

The expansion of the first term in (\ref{eq3}) yields:
\begin{equation}
\label{eq4}
\begin{aligned}
I\left( Y,\hat{X} \right)= & \int p(y, \hat{x}) \log \frac{p(y, \hat{x})}{p(y) p(\hat{x})} d y d \hat{x} \\
= & \underbrace{- \int{p}(y,\hat{x})\log p(y)dyd\hat{x}}_{H\left( Y \right)=\text{constant}}\\
& + \int{p}(y,\hat{x})\log p(y\mid \hat{x})dyd\hat{x},
\end{aligned} 
\end{equation}
where the first term of the second equation is the entropy of $Y$. For a given graph and a given task, the entropy of $Y$ is determined, so this term can be regarded as a constant and can be ignored in subsequent optimization. Therefore, we can obtain the expression for (\ref{eq4}) in the optimization implication:
\begin{equation}
\label{eq5}
    I\left( Y,\hat{X} \right) = \int{p}(y,\hat{x})\log p(y\mid \hat{x})dyd\hat{x}.
\end{equation}

The joint distribution $p\left( g,y \right)$ for graph data and target labels is known. ${{p}_{\phi }}\left( \hat{x}|g \right)$ is determined by the transmitter network ${{p}_{\phi }}\left( x|g \right)$ and channel model ${{p}_{channel}}\left( \hat{x}|x;\epsilon  \right)$. For the second integral term, we derive according to the Markov chain:
\begin{equation}
\label{eq6}
    p(y \mid \hat{x}) =\int \frac{p(g,y) {{p}_{\phi }}(\hat{x} \mid g)}{p(\hat{x})} d g,
\end{equation}

Next, we address the second mutual information term of (\ref{eq3}) and expand it:
\begin{equation}
\label{eq7}
    I\left( G,\hat{X} \right)=\int{p}(\hat{x}\mid g)p(g)\log \frac{{{p}_{\phi }}(\hat{x}\mid g)}{p(\hat{x})}dgd\hat{x},
\end{equation}
where $p(\hat{x})$ is also an intractable high-dimensional integral:
\begin{equation}
\label{eq8}
    p(\hat{x}) =\int p(g) {{p}_{\phi }}(\hat{x} \mid g) d g. 
\end{equation}
It is customary in IB to substitute this marginal distribution with a tractable prior distribution\cite{DBLP:conf/iclr/AlemiFD017}. In GIB, finding a suitable variational prior is difficult. This is because of GIB's interpretation of $p(\hat{x})$: it represents the distribution of irregular subgraph structures, not just a latent graph data representation. Moreover, due to the non-IID nature of graph data, finding a simple function as a prior distribution is not feasible.
Therefore, a new method is needed to estimate that mutual information.

\item \textbf{Problem 2}: How to utilize topological information in graph data for communication tasks?

By designing a feature extractor  based on GIB theory, it becomes possible to selectively extract task relevant information, reducing redundancy and improving efficiency. Yet, when GNNs extract features from graph data, they often concentrate too much on nodes and neglect the graph's structural features.
However, for many tasks, the crucial information is deeply rooted in the graph's topology. Therefore, in the process of feature extraction, it is necessary to consider the topological structure of the graph. 
To do this, we need to impose constraints on the feature extractor to improve the role of topological information in feature extraction. 

\item \textbf{Problem 3}: How to be compatible with digital communication systems?

While the proposed task-oriented communication system effectively addresses the challenge of graph data transmission, its use of continuous signals poses compatibility issues with existing digital communication systems.  Given the well-established infrastructure of digital communication systems, it is impractical to abandon it completely.  Therefore, it is necessary to develop a task-oriented communication system compatible with the digital communication system to transmit graph data.

\end{itemize}


\section{GIB-Enabled Task-oriented Communication}

In this section, we develop a task-oriented communication system based on GIB theory. Specifically, we combine variational approximation and MINE techniques to improve the GIB formulation \cite{Belghazi2018MutualIN}.
This modification enables a more streamlined optimization process employing neural networks, which solves \textbf{Problem 1}. In addition, we describe the process of feature extraction in detail and also node assignment, including the network configuration and the pertinent revision made to the objective function, which solves \textbf{Problem 2}.

\subsection{Graph Information Bottleneck Reformulation}
To solve \textbf{Problem 1},
we introduce the variable distribution ${{q}_{\theta }}\left( y|\hat{x} \right)$ as a substitute for the true posterior distribution $p\left( y|\hat{x} \right)$. $\theta $ represents the parameter of the inference neural network in the receiver computing the inference output $\hat{y}$. Consequently, (\ref{eq5}) is transformed into:
\begin{equation}
\begin{aligned}
\label{eq9}
I\left( Y,\hat{X} \right) & = \int{p}\left( y,\hat{x} \right)\log {{q}_{\theta }}\left( y|\hat{x} \right)dyd\hat{x}\\
& ={{E}_{p\left( y,\hat{x} \right)}}\left[ \log {{q}_{\theta }}\left( y|\hat{x} \right) \right].
\end{aligned}
\end{equation}
Subsequently, Monte Carlo sampling is used to derive the empirical distribution of the joint distribution as an approximation. $N$ samples are taken for the given data, that is:
\begin{equation}
\label{eq10}
    p\left( y,\hat{x} \right)\approx \frac{1}{N}\sum\limits_{i=1}^{N}{{{\delta }_{y}}}\left( {{y}_{i}} \right){{\delta }_{{\hat{x}}}}\left( {{{\hat{x}}}_{i}} \right).
\end{equation}
Here, $\delta \left( \cdot  \right)$ denotes the Dirac function utilized for sampling the training data. ${{y}_{i}}$ and ${{\hat{x}}_{i}}$ represent the label and representation received by the receiver for the $i$-th training data, respectively. This approximation leads to a tractable optimization objective from (\ref{eq9}):
\begin{equation}
\label{eq11}
{{\mathcal{L}}_{inf}}\left( {{q}_{\theta }}\left( y\mid \hat{x} \right) \right) = -\frac{1}{N}\sum\limits_{i=1}^{N}{\log }{{q}_{\theta }}\left( {{y}_{i}}\mid {{{\hat{x}}}_{i}} \right).
\end{equation}

Minimizing the objective function in (\ref{eq3}) is consistent with maximizing $I\left( Y,\hat{X} \right)$ and minimizing ${\mathcal{L}}_{inf}$.
This function is defined as the loss between the inference result $Y$ and the ground truth for the received $\hat{x}$. A smaller loss signifies superior performance of the inference neural network.

\begin{figure}
    \centering
    \includegraphics[width=0.45\textwidth]{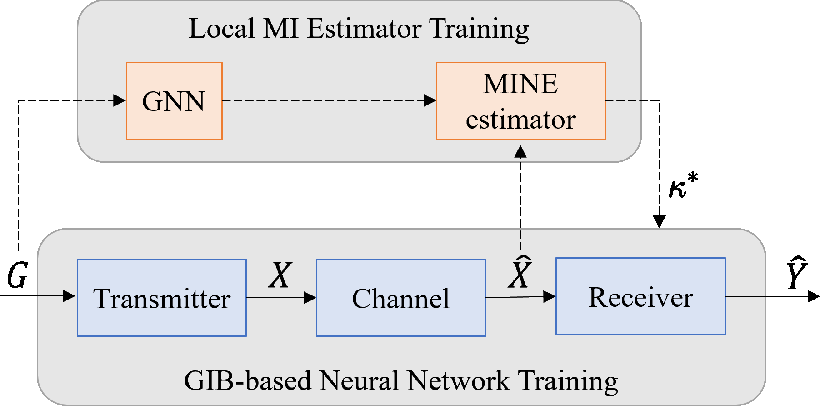}
    \caption{
    Training Framework for GIB-Based Task-Oriented Communication Systems: The process begins with training the mutual information estimator to acquire a set of optimized MINE parameters, denoted as 
${{\kappa }^{*}}$. Following this, the entire neural network undergoes training, during which the mutual information estimator utilizes the previously determined MINE parameters.}
    \label{fig2}
\end{figure}

About the second mutual information term $I\left( G,\hat{X} \right)$ of (\ref{eq3}), indicating the transmission informativeness, we employ the MINE to directly approximate it without the need to estimate $p(\hat{x})$. We express this term in the form of Kullback-Leibler (KL)-divergence:
\begin{equation}
\label{eq12}
    I\left( G,\hat{X} \right)={{D}_{KL}}\left[ P\left( G,\hat{X} \right)||P\left( G \right)\otimes P\left( {\hat{X}} \right) \right].
\end{equation}
For ease of analysis and mathematical treatment, we adopt the Donsker-Varadhan representation of KL-divergence, which expresses the mutual information term as the difference between the expected value and the logarithmic expected value:

\begin{equation}
\begin{aligned}
   \label{eq13}
    I\left( G,\hat{X} \right)&= \underset{T:G\times \hat{X}\to \mathbb{R}}{\mathop{\sup }}\,{{E}_{P\left( G,\hat{X} \right)}}\left[ T \right]-\log {{E}_{P\left( G \right)P\left( {\hat{X}} \right)}}\left[ {{e}^{T}} \right]\\
    &= {{\mathcal{L}}_{MI}}\left( T \right),     
\end{aligned}
\end{equation}
in which $T={{f}_{\kappa }}\left( g,\hat{x} \right)$ encompasses all functions that render two expectations finite. The MINE can be used to estimate the mutual information between regular input data and its vector representation. Due to the irregularity of graph data, GNN is employed to extract the vector representation before feeding it into the multi-layer perceptron (MLP) for processing alongside the representation $\hat{X}$. ${{f}_{\kappa }}\left( \cdot  \right)$ serves as the statistics network, denoting the neural network that executes the process of obtaining the corresponding real numbers from $G$ and $\hat{X}$. The optimization of this estimator involves adjusting the MINE parameter $\kappa$ such that the value on the right-hand side of (\ref{eq12}) closely approximates $I\left( G,\hat{X} \right)$. This optimization is formalized as:
\begin{equation}
\label{eq14}
    \underset{\kappa }{\mathop{\max }}\, {{\mathcal{L}}_{MI}}\left( \kappa ,\hat{X} \right)=\frac{1}{K}\sum\limits_{i=1}^{K}{{{f}_{\kappa }}}\left( {{g}_{i}},{{{\hat{x}}}_{i}} \right)-\log \frac{1}{K}\sum\limits_{i=1,j\ne i}^{K}{{{e}^{{{f}_{\kappa }}\left( {{g}_{i}},{{{\hat{x}}}_{i}} \right)}}}.
\end{equation}
After training with $K$ sets of training data, the training process generates a set of suboptimal MINE parameters denoted as ${{\kappa }^{*}}$ (see Fig. \ref{fig2}).

Drawing from (\ref{eq9}) and (\ref{eq13}), we derive a reformulation of the overall loss function:
\begin{equation}
\begin{aligned}
\label{eq15}
      {{\mathcal{L}}_{GIB}}&\left( \phi ,\theta ,{{\kappa }^{*}} \right)= {{E}_{p\left( g,y \right)}}\Big\{{{E}_{{{p}_{\phi }}\left( \hat{x}|g \right)}}\left[ -\log {{q}_{\theta }}\left( y|\hat{x} \right) \right]\\
    &+\beta \left[ {{E}_{{{p}_{\phi }}\left( \hat{x}|g \right)}}{{f}_{{{\kappa }^{*}}}}\left( g,\hat{x} \right)-\log {{E}_{p\left( {\hat{x}} \right)}}{{e}^{{{f}_{{{\kappa }^{*}}}}\left( g,\hat{x} \right)}} \right] \Big\}  .
\end{aligned}
\end{equation}
By combining the sampling method, equations (\ref{eq11}) and (\ref{eq14}), we arrive at a tractable optimization problem for the entire system:
\begin{equation}
\label{eq16}
    \underset{\phi ,\theta }{\mathop{\min }}\,{{\mathcal{L}}_{GIB}}\left( \phi ,\theta ,{{\kappa }^{*}} \right)={{\mathcal{L}}_{inf}}\left( \theta,\phi \right)+\beta {{\mathcal{L}}_{MI}}\left( {{\kappa }^{*}},\hat{X} \right).
\end{equation}

\subsection{Model Training Strategy}
To solve \textbf{Problem 2}, we specially design the feature extraction module. Node assignment is the main step in the graph feature extraction process. First, we introduce assignment probabilities as a continuous relaxation in node assignment, to address the problem that the discrete nature of the graph make it difficult to optimize (\ref{eq16}) using gradient optimization methods.
Next, we introduce connectivity loss to improve the impact of topological information on feature extraction, which can also make the node assignment process more stable. 
The detailed design process of this module will be discussed below.

We devise a node assignment mechanism to determine the inclusion of each node in the graph within the subgraph. This mechanism involves GNN extracting node features $X$ from the original graph $G$. These features are then input into an MLP to obtain preliminary node assignments. The $\rm Softmax$ function is subsequently employed to convert these assignments into a probabilistic form. We represent the selected subgraph as ${{G}_{sub}}$. Specifically, for a given graph $G$, nodes either belong to ${{G}_{sub}}$ or ${{\overline{G}}_{sub}}$. The node assignment mechanism yields a matrix $S$:
\begin{equation}
\label{eq17}
    S={\rm Softmax} \left( {{MLP}_{{{\sigma}_{2}}}}\left( X \right) \right) \; with \; X={{GNN}_{{{\sigma}_{1}}}}\left( A,X \right).
\end{equation}
The matrix $S$ has dimensions $m\times 2$, where $m$ is the number of nodes in the input graph, and each row is a two-dimensional vector indicating the assignment probability of the corresponding node. Specifically, the $i$-th row of $S$,
\[ \left[ p\left( {V_{i}}\in {{G}_{sub}}\mid {V_{i}} \right),p\left( {V_{i}}\in {{\overline{G}}_{sub}}\mid {V_{i}} \right) \right], \]
represents the probability of the $i$-th node belonging to ${{G}_{sub}}$ or ${{\overline{G}}_{sub}}$.

We define ${{\phi }_{1}}=\left\{ {{\sigma }_{1}},{{\sigma }_{2}} \right\}$ as the parameter of the node assignment mechanism, which is part of the parameter of the transmitter. Once adequately trained, the values in each row of $S$ should converge to 0 or 1, achieving robust node assignment. We take the first row of ${{S}^{T}}X$ to obtain the feature vectors of the $n$ nodes belonging to the subgraph. 

In conjunction with the JSC encoder parameter ${{\phi }_{2}}$, we obtain the transmitter parameter $\phi =\left\{ {{\phi }_{1}},{{\phi }_{2}} \right\}$. For simplicity, we collectively refer to the node assignment mechanism and the JSC encoder as the feature extractor and encoder, denoted as ${{g}_{\phi }}\left( \cdot \right)$. Consequently, the objective function is explicitly expressed as:
\begin{equation}
\label{eq18}
    \underset{\phi ,\theta }{\mathop{\min }}\,{{\mathcal{L}}_{GIB}}\left( \phi ,\theta ,{{\kappa }^{*}} \right)={{\mathcal{L}}_{inf}}\left( {{q}_{\theta }}\left( {{g}_{\phi }}\left( G \right) \right) \right)+\beta {{\mathcal{L}}_{MI}}\left( {{\kappa }^{*}},\hat{X} \right).
\end{equation}

We introduce a continuous relaxation with probabilistic assignment of nodes, alleviating the problems arising from the discreteness of the graph. Nevertheless, inadequate initialization may lead to poorly trained node allocation mechanisms and failure to achieve the desired outcomes. In other words, if the probabilities of a node belonging to ${{G}_{sub}}$ or not are too close, nodes amay not be appropriately assigned. On one hand, over-assigning nodes to ${{G}_{sub}}$ will lead to the presentation of a subgraph that includes an excessive amount of redundant information. On the other hand, assigning too few nodes to ${{G}_{sub}}$ will result in an inadequate amount of task-related information in the subgraph, rendering it incapable of successfully executing the task.

To address the above issues, we assume that the model has an inductive bias that helps the model to focus more on the connectivity relationships between nodes. The model is thus better able to capture and utilize the topological information of the graph. We incorporate the connectivity loss proposed in \cite{yu2021graph} to introduce this inductive bias:
\begin{equation}
\label{eq19}
    {{\mathcal{L}}_{con}}={{\left\| Norm\left( {{S}^{T}}AS \right)-{{I}_{2}} \right\|}_{F}},
\end{equation}
where $Norm\left( \cdot  \right)$ denotes row normalization, ${{I}_{2}}$ is the $2\times 2$ identity matrix , and ${{\left\| \cdot  \right\|}_{F}}$ is the Frobenius norm. Elements ${{a}_{11}}$ and ${{a}_{12}}$ in the first row of ${{S}^{T}}AS$ are defined as follows:
\begin{equation}
\label{eq20}
\begin{aligned}
     {{a}_{11}}=\sum\limits_{i,j}{{{A}_{ij}}}p\left( {V_{i}}\in {{G}_{sub}}\mid {V_{i}} \right)p\left( {V_{j}}\in {{G}_{sub}}\mid {V_{j}} \right),
\end{aligned}
\end{equation}
and 
\begin{equation}
\label{eq21}
\begin{aligned}
    {{a}_{12}}=\sum\limits_{i,j}{{{A}_{ij}}}p\left( {V_{i}}\in {{G}_{sub}}\mid {{N}_{i}} \right)p\left( {V_{j}}\in {{\overline{G}}_{sub}}\mid {V_{j}} \right).
\end{aligned}
\end{equation}

Intuitively, if a node belongs to ${{G}_{sub}}$, its neighboring nodes are highly probable to belong to ${{G}_{sub}}$ as well. Conversely, if a node does not belong to ${{G}_{sub}}$, its adjacent nodes likely do not belong to ${{G}_{sub}}$ either. Consistently with this, ensuring adequate nodes are assigned to ${{G}_{sub}}$ while reducing redundancy is achieved through $\frac{{{a}_{11}}}{{{a}_{11}}+{{a}_{12}}}\to 1$. This occurs simultaneously with reducing redundancy in ${{G}_{sub}}$ through $\frac{{{a}_{12}}}{{{a}_{11}}+{{a}_{12}}}\to 0$. Analogously, this holds for ${{\overline{G}}_{sub}}$ and the elements of the second row of ${{S}^{T}}AS$.

\begin{algorithm}[tb]
   \caption{Model Training Procedure}
   \label{algorithm}
\begin{algorithmic}[1]
   \STATE {\bfseries Input:} Graph and class label pairs $\{G, Y\}$, dimension of the encoder output $D$, batch size $B$, channel variance $\frac{N_0}{2}$
   \WHILE{epoch t=1 to $T$}
        \STATE Select a mini-batch of data ${\{(g_b, y_b)\}}_{b=1}^B$
        \STATE Extract node feature vectors 
        \STATE Compute the node assignment matrix $S$
        \STATE Compute the encoded feature vectors of subgraphs $X$
        \WHILE{batch $b$=1 to $B$}
            \STATE Sample noise $\epsilon$ $N$ times for each pair of $(g_b, y_b)$
        \ENDWHILE
        \STATE Compute the connectivity loss based on (\ref{eq19})
        \STATE Compute the inference loss based on (\ref{eq11})
        \WHILE{inner loop $k=1$ to $K$}
            \STATE Tanin the MINE by (\ref{eq14}) to get the suboptimal ${{\kappa }^{*}}$
        \ENDWHILE
        \STATE Compute the MI loss with ${{\kappa }^{*}}$:
        
        ${{\mathcal{L}}_{MI}}=\frac{1}{N}\sum\limits_{i=1}^{N}{{f}_{{\kappa }^{*} }}\left( {{g}_{i}},{{{\hat{x}}}_{i}} \right)-\log \frac{1}{N}\sum\limits_{i=1,j\ne i}^{N}{{e}^{{{f}_{{\kappa }^{*}}}\left( {{g}_{i}},{{{\hat{x}}}_{i}} \right)}}$
        
        \STATE Compute the overall loss ${\mathcal{L}}_{GIB}$ based on (\ref{eq15})
        \STATE Calculate gradients and update model parameters by backpropagation
   \ENDWHILE
\end{algorithmic}
\end{algorithm}

To summarize, we introduce the continuous relaxation with probabilistic node assignment to enhance the optimization process. To address potential challenges stemming from inadequate initialization, such as ambiguous node assignment and unstable training process, we incorporate the inductive bias with the connectivity loss ${{\mathcal{L}}_{con}}$. The overall loss function is refined as:

\begin{equation}
\label{eq22}
\begin{aligned}
      \underset{\phi ,\theta }{\mathop{\min }}\,{{\mathcal{L}}_{GIB}}\left( \phi ,\theta ,{{\kappa }^{*}} \right)={{\mathcal{L}}_{inf}}\left( {{q}_{\theta }}\left( {{g}_{\phi }}\left( G \right) \right) \right)\\
      +\beta {{\mathcal{L}}_{MI}}\left( {{\kappa }^{*}},\hat{X} \right)+\alpha {{\mathcal{L}}_{con}}\left( {{g}_{\phi }}\left( G \right) \right).  
\end{aligned}
\end{equation}

The training procedures for the GIB-enabled task-oriented communication system are illustrated in Algorithm \ref{algorithm}.

\subsection{Computational Complexity Analysis}
The computational complexity inherent in the proposed methodology primarily emanates from its reliance on GNNs. In our analysis, the GCN is chosen as the representative GNN to elucidate the computational complexity of the system. Characteristically, GCN employs a one-hop receptive field to assimilate local features, subsequently enlarging this receptive field through the stratification of layers. Denoting the number of GCN layers by $L$, the computational complexity attributed to GNNs can be articulated as $\mathcal{O}\left(|E|\sum\limits_{i=1}^{L}{D_{in}^{(i)}D_{out}^{(i)}}\right)$, where $|E|$ signifies the quantity of edges in the graph, and $D_{in}^{(i)}$, $D_{out}^{(i)}$ represent the input and output dimensions of each respective layer. In parallel, the complexity associated with the MLP layer and the $\rm Softmax$ function within the node assignment mechanism is quantified as $\mathcal{O}(mD)$ and $\mathcal{O}(m)$ respectively, with 
$m$ indicating the total number of graph nodes and 
$D$ symbolizing the output dimension of GNNs. Consequently, the cumulative computational complexity of the proposed framework can be comprehensively expressed as $\mathcal{O}(|E|\sum\limits_{i=1}^{L}{D_{in}^{(i)}D_{out}^{(i)}}+ND)$.

\section{Digitization of GIB-Enabled Task-Oriented Communications}

This section extends the analog task-oriented communication system, introduced in Section \uppercase\expandafter{\romannumeral3}, to suit digital communication environments. As outlined in \textbf{Problem 3}, practical applications necessitate compatibility with digital systems. We address this by integrating vector quantization for digital transmission adaptation.

\subsection{Digital Transmission}
A pivotal strategy for transitioning to a digital framework involves discrete codebook mapping. This technique maps the continuous outputs of the neural network encoder into discrete codewords, as discussed in related works \cite{Xie2022RobustIB,Bo2022LearningBJ}. The mapping process employs a predefined discrete codebook alongside nearest neighbor rules \cite{Hu2022RobustSC}. Let ${{x}_{1}},{{x}_{2}},\ldots ,{{x}_{n}}\in \mathbb{R}^d$ be the continuous outputs from the encoder described in Section III. The codebook, denoted as $\mathcal{E}\in {\mathbb{R}^{K\times d}}$ consists of $K$ codewords ${{e}_{1}},{{e}_{2}},\ldots ,{{e}_{K}}\in\mathbb{R}^{d}$, where $K$ represents the codebook size. The discrete mapping function is formulated as:
\begin{equation}
\label{eq23}
    {{x}_{i\_q}}={{e}_{k}}\text{ where }k=\arg \underset{j}{\mathop{\min }}\,{{\left\| {{x}_{i}}-{{e}_{j}} \right\|}_{2}}.
\end{equation}
Employing discrete codebook mapping, the probabilistic model of the system is transformed into a new Markov chain, represented as:
\begin{equation}
\label{eq24}
\left( Y \right)G \to X \to Z \to \hat Z \to \hat X \to \hat Y,
\end{equation}
which adheres to the following relationship:
\begin{equation}
\label{eq25}
p\left( \hat{y}|g \right)={{p}_{\phi }}\left( x|g \right){{p}_{Q}}\left( z|x \right){{p}_{SDC}}\left( \hat{z}|z \right){{p}_{DQ}}\left( \hat{x}|\hat{z} \right){{p}_{\inf }}\left( \hat{y}|\hat{x} \right).
\end{equation}
Here, the transition from 
 $X$ to $Z$ entails discretizing feature representations using the codebook. 
 $Z$ represents the index sequence derived via the nearest neighbor rule, substituting $X$  as the input to the channel.
 For the discrete signal $Z$, we employ the discrete channel defined in (\ref{unifiedChannel}) for its transmission, that is:
 \begin{equation}
     {{p}_{SDC}}\left( \hat{z}|z \right)=p_{\text{channel}}(\hat{z}|z; \zeta=1)
     =P_{\hat{Z}|Z}(\hat{z}|z),
 \end{equation}
 where $P_{\hat{Z}|Z}(\hat{z}|z)$ is the transition probability matrix of SDC described in Section II when $\zeta=1$.
 \\

\begin{figure}
    \centering
    \includegraphics[width=0.5\textwidth]{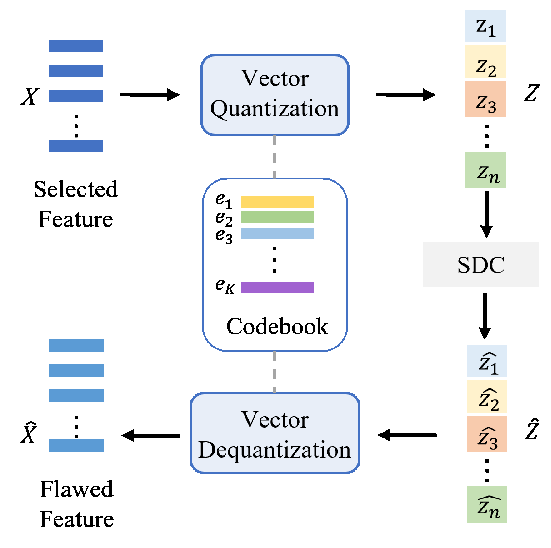}
    \caption{System Digitization Modules. This figure illustrates the process where the subgraph representation is matched against a codebook to generate an index sequence via the nearest neighbor principle. This sequence is sent through a symmetric discrete channel, subject to a specified error probability, leading to potential discrepancies between the transmitted and received index sequences. The receiver then reconstructs the representation vector using a shared codebook with the transmitter for subsequent task inference. }
    \label{fig3}
\end{figure}

In digital transmission, as indicated in (\ref{eq23}), each vector $x_i$ from the encoder output is aligned with its nearest embedding vector $e_k$ from the shared codebook. The transmitter's role is simplified to sending the index $k$ to the receiver. The assignment of indices for transmission is defined as:
\begin{equation}
\label{eq26}
    p({{Z}_{i}}=k\mid {{x}_{i}})=\left\{ \begin{array}{*{35}{l}}
   1 & \text{ for }k=\arg \underset{j}{\mathop{\min }}\,{{\left\| {{x}_{i}}-{{e}_{j}} \right\|}_{2}}  \\
   0 & \text{ otherwise }  \\
\end{array} \right..
\end{equation}
This results in a continuous vector being translated into a discrete one-hot codeword. However, channel imperfections can lead to erroneous index detection ($\hat{Z}$). We utilize the SDC to model this transmission of discrete indices.

\subsection{Design of Discrete Codebook}
Each codeword $e_i$ in the vector quantizer's codebook defines a Voronoi region ${\mathcal{V}}_{i}$:
\begin{equation}
\label{eq28}
    {{\mathcal{V}}_{i}}=\left\{ x\in {{R}^{d}}:\left\| x-{{e}_{i}} \right\|\le \left\| x-{{e}_{j}} \right\|,\text{ for all }j\ne i \right\}.
\end{equation}
These regions collectively span the entire vector space $\mathbb{R}^d$
  to which the encoder's output belongs. The quantization process entails finding the codeword closest to a given vector, which subsequently determines the Voronoi region and index for the vector. Balancing the size of the codebook is crucial as increasing it reduces quantization distortion but also elevates computational complexity.

We opted for a moderate-sized, suboptimally designed codebook to achieve acceptable quantization results. To enhance the codebook's performance, we introduce a learnable aspect, allowing codewords to adapt during training. Additionally, the encoder's output range is constrained to prevent extreme variations.

The non-differentiable nature of the quantization operation poses a challenge for encoder training. To address this, a straight-through estimator is employed, allowing gradients from the decoder input to flow back to the encoder output. Furthermore, a loss term ($\mathcal{L}_{vq}$) is added to reduce the distance between the encoder output and the corresponding codeword:
\begin{equation}
\label{eq29}
    {{\mathcal{L}}_{vq}}={{\left\| sg\left[ x \right]-e \right\|}_{2}},
\end{equation}
with $sg\left[ \cdot  \right]$ representing a stop gradient function. 

A commitment loss ($\mathcal{L}_{CM}$) is also introduced to ensure the encoder's output does not deviate excessively from the codewords:
\begin{equation}
\label{eq30}
    {{\mathcal{L}}_{CM}}={{\left\| x-sg\left[ e \right] \right\|}_{2}}.
\end{equation}

The final objective function (
$\mathcal{L}_{VQ-GIB}$) encompasses these elements alongside the mutual information and connectivity losses, guiding the entire system:
\begin{equation}
\label{eq31}
\begin{aligned}
    {{\mathcal{L}}_{VQ-GIB}}&={{\mathcal{L}}_{inf}}\left( {{q}_{\theta }}\left( {{g}_{\phi }}\left( G \right) \right) \right)+\beta {{\mathcal{L}}_{MI}}\left( {{\kappa }^{*}},\hat{X} \right)\\
    &\quad+\alpha {{\mathcal{L}}_{con}}\left( {{g}_{\phi }}\left( G \right) \right)+{{\mathcal{L}}_{vq}}+\lambda {{\mathcal{L}}_{CM}}
\end{aligned}
\end{equation}
subject to the optimized mutual information parameters $\quad {{\kappa }^{*}}$:
\begin{equation}
\label{eq32}
\quad {{\kappa }^{*}}=\underset{\kappa }{\mathop{\arg \max }}\,{{\mathcal{L}}_{MI}}\left( \kappa ,\hat{X} \right).
\end{equation}

\begin{equation}
\label{eq33}
    {{\mathcal{L}}_{vq}}={{\left\| sg\left[ x \right]-e \right\|}_{2}},
\end{equation}
where $sg\left[ \cdot  \right]$ denotes stop gradient. Therefore, this item is solely valid for codebook learning, not encoder training.

Given that the volume of the encoder's output space is dimensionless, achieving satisfactory quantization mappings is challenging if the training of the codewords in the codebook does not align with the rate of parameter training in the encoder. Therefore a commitment loss is introduced to constrain the encoder's output from increasing:
\begin{equation}
\label{eq34}
    {{\mathcal{L}}_{CM}}={{\left\| x-sg\left[ e \right] \right\|}_{2}}.
\end{equation}
This item only affects the encoder, bringing the output closer to the code vector without influencing codebook learning.

For codebook updating, we adopt an exponential moving average (EMA) approach akin to $K$-means clustering. This method assigns variable weights to cluster centers over training batches, ensuring that the codebook dynamically adapts to the encoder's output.

\section{Experiments and Discussions}

In this section, we evaluate the performance of the proposed GIB-based task-oriented communication system on graph classification tasks, and investigate the adaptability and effectiveness of continuous and discrete communication systems respectively. 
Additionally, ablation studies are also conducted to illustrate the contributions of the MI loss $\mathcal{L}_{MI}$ in GIB and the connectivity loss $\mathcal{L}_{con}$ presented in Section \uppercase\expandafter{\romannumeral3}, as well as the impact of the trade-off factor $\beta$ on the system performance.


\subsection{Experimental Setup}
\subsubsection{Datasets}

For our graph classification experiments, we carefully select two datasets: COLLAB and PROTEINS.

\begin{itemize}
    \item \textbf{COLLAB}: This scientific collaboration dataset represents a researcher's ego network, where nodes correspond to researchers and edges indicate collaboration between them. Each researcher's ego network is labeled based on the field to which the researcher belongs, resulting in three possible labels. COLLAB consists of 5,000 graphs, with an average of 74 nodes and 2,457 edges per graph.
    \item \textbf{PROTEINS}: This dataset comprises 1,113 proteins classified as enzymes or non-enzymes. Nodes in the graph represent amino acids, and edges exist between nodes if the distance between corresponding amino acids is less than 6 angstroms. On average, each graph in the PROTEINS dataset has 39 nodes and 73 edges.
\end{itemize}

\begin{table}[t]
\centering
\caption{Architecture of Neural Network for GIB-Enabled Task-Oriented Communications}
\label{tab1}
\resizebox{\linewidth}{!}{
\begin{tabular}{c|c|c|c}
\hline
  & \multicolumn{2}{c|}{\textbf{Layer}}&\textbf{\makecell[c]{Output\\Dimensions}}\\
\hline
\multirow{5}{*}[-0.3cm]{\makecell[c]{Encoder\\Network}} & \multicolumn{2}{c|}{GCN Layer} & ($\mathrm{num}$, $\mathrm{dim}$)\\
\cline{2-4}
  & \multicolumn{2}{c|}{GCN Layer} & ($\mathrm{num}$, $\mathrm{dim}$) \\
\cline{2-4}
  & \multirow{2}*{\makecell[c]{Node\\Assignment}} & \makecell[c]{Fully-connected Layer\\ + Tanh} & ($\mathrm{num}$, $\mathrm{dim}$) \\
\cline{3-4}
  & & \makecell[c]{Fully-connected Layer\\ + $\rm Softmax$} & ($\mathrm{num}$, 2) \\
\cline{2-4}
  & \multicolumn{2}{c|}{Aggregation} & ($\mathrm{batch}$, $\mathrm{dim}$) \\
\hline
Channel & \multicolumn{2}{c}{AWGN Channel} & ($\mathrm{batch}$, $\mathrm{dim}$) \\ \hline
\multirow{2}*{\makecell[c]{Decoder\\Network}} & \multicolumn{2}{c|}{Fully-connected Layer + Dropout} & ($\mathrm{batch}$, $\mathrm{dim}$)\\
\cline{2-4}
  &\multicolumn{2}{c|}{Fully-connected Layer + Log-softmax} & ($\mathrm{batch}$, $\mathrm{class\_n}$) \\
\hline
\end{tabular}}
\end{table}

\begin{table}[t]
\centering
\caption{Architecture of Neural Network for GIB-Enabled Discrete Task-Oriented Communications.}
\label{tab2}
\resizebox{\linewidth}{!}{
\begin{tabular}{c|c|c|c}
\hline
  & \multicolumn{2}{c|}{\textbf{Layer}}&\textbf{\makecell[c]{Output\\Dimensions}}\\
\hline 
\multirow{5}{*}[-0.1cm]{\makecell[c]{Encoder\\Network}} & \multicolumn{2}{c|}{GCN Layer} & ($\mathrm{num}$, $\mathrm{dim}$)\\  
\cline{2-4}
  & \multicolumn{2}{c|}{GCN Layer} & ($\mathrm{num}$, $\mathrm{dim}$) \\
\cline{2-4}
  & \multicolumn{2}{c|}{Node Assignment} & ($\mathrm{batch}$, $\mathrm{dim}$) \\
\cline{2-4}
  & \multicolumn{2}{c|}{Aggregation} & ($\mathrm{batch}$, $\mathrm{dim}$) \\
  \cline{2-4}
  & \multicolumn{2}{c|}{Vector Quantization} & ($\mathrm{batch}$, $\mathrm{size}$) \\
\hline

Channel & \multicolumn{2}{c}{Symmetric Discrete Channel} & ($\mathrm{batch}$, $\mathrm{size}$) \\ \hline

\multirow{3}*{\makecell[c]{Decoder\\Network}} & \multicolumn{2}{c|}{Vector Decoding} & ($\mathrm{batch}$, $\mathrm{dim}$)\\
\cline{2-4}
  & \multicolumn{2}{c|}{Fully-connected Layer + Dropout} & ($\mathrm{batch}$, $\mathrm{dim}$)\\
\cline{2-4}
  &\multicolumn{2}{c|}{Fully-connected Layer + Log-softmax} & ($\mathrm{batch}$, $\mathrm{class\_n}$) \\
\hline
\end{tabular}}
\end{table}

\subsubsection{Setings and Baselines}

To assess the performance of the proposed method in a graph classification task, we integrate GIB into two distinct backbones: Graph Convolutional Network (GCN) \cite{DBLP:conf/iclr/SunHV020} and Graph Isomorphism Network (GIN) \cite{DBLP:conf/iclr/XuHLJ19}.

We compare the proposed method with other graph-level representation learning methods: \textbf{InfoGraph} based on mean aggregation \cite{DBLP:conf/iclr/SunHV020} and \textbf{ASAP} (Adaptive Structure Aware Pooling) based on pooling aggregation \cite{Ranjan2019ASAPAS} in terms of graph classification accuracy, respectively. 

\begin{itemize}
    \item \textbf{InfoGraph}: InfoGraph is an unsupervised graph-level representation learning method that maximizes the mutual information between the representations of entire graphs and the representations of substructures at different granularity (e.g., nodes, edges, triangles) to make the graph representations adequately capture the features of substructures.
    \item \textbf{ASAP}: ASAP utilizes self-attention network along with GNNs to capture the importance of each node in a given graph. It also learns a sparse soft cluster assignment for nodes at each layer to pool the subgraphs to form the pooled graph.
\end{itemize}

For a fair comparison, all methods use the same number of GNN layers in the backbone. Models are trained using Stochastic Gradient Descent (SGD) with the Adam optimizer. We employ 10-fold cross-validation to report classification accuracy in experiments to validate the performance of the models.

\subsubsection{Neural Network Architecture}

 While we do not enforce complete consistency in the network architectures of different approaches, we ensure that the GNN backbone and the number of layers are consistent across methods. The neural network architectures for the proposed method are shown in Tables \ref{tab1} and \ref{tab2}, which are outlined below. In the tables, $\mathrm{num}$ is the number of nodes, $\mathrm{batch}$ is the batch size during training, $\mathrm{class\_n}$ is the number of graph classes, and $\mathrm{dim}$ represents the hidden dimension. The $\mathrm{size}$ in Table \ref{tab2} refers to the size of the codebook, which was set to 256 in the experiment.

\begin{itemize}
    \item Table \ref{tab1} shows the network architecture of the proposed method where GCN is used as the backbone. We use two GCN layers to extract the node features of the graph and then perform node assignment. The output of the Node Assignment layer is multiplied with the full node features extracted by the GCN to obtain the node features of the selected subgraph. This output undergoes power normalization before passing through the AWGN channel.
    \item Table \ref{tab2} illustrates the network architecture of the proposed digital communication system.  In discrete communication systems, subgraph representations require quantization before entering the channel. During the training process, the codebook utilized for vector quantization undergoes automatic updates until it reaches a stabilized state. The index sequence obtained from quantization passes through a symmetric discrete channel with a certain error transfer probability, and the decoder performs dequantization based on the codebook shared with the encoder.
\end{itemize}

We note that the hidden layer dimensions of the proposed method and baseline methods are set to the same. To comply with the original implementation idea of the baseline methods, the output dimension of their encoder network is twice as large as that of the proposed method, consuming more communication resources. In addition, since the vector quantization part is not involved in baselines, the same quantization method as the proposed method is adopted for comparison algorithms.

\subsection{Performance of GIB-enabled Task-oriented Communications Without Vector Quantization}

\begin{figure*}[t]
    \centering
    \begin{subfigure}[b]{0.4\textwidth}
    \includegraphics[width=\textwidth]{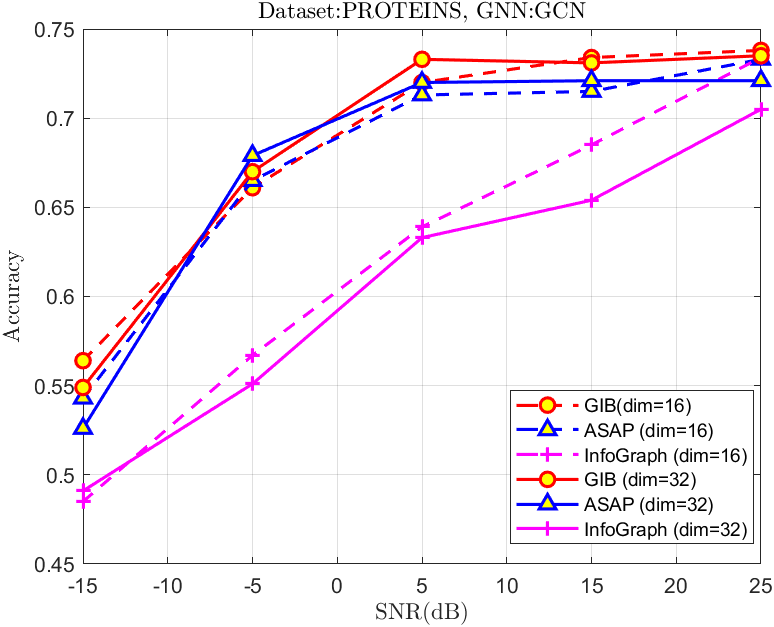}
    \captionsetup{justification=centering}
    \caption{PROTEINS}
    \end{subfigure}
    \hspace{0.1\textwidth}
    \begin{subfigure}[b]{0.4\textwidth}
    \includegraphics[width=\textwidth]{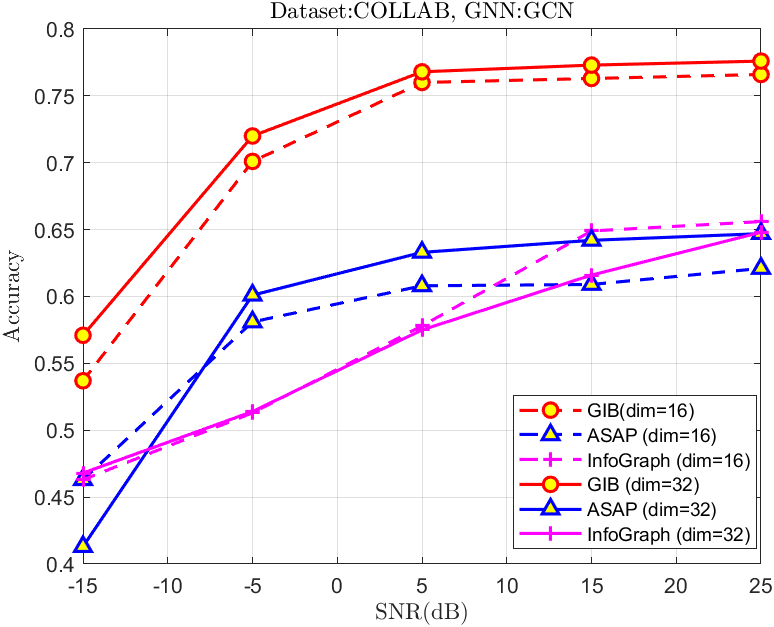}
    \captionsetup{justification=centering}
    \caption{COLLAB}
    \end{subfigure}
    \caption{Classification accuracy variation with SNR for three methods on PROTEINS and COLLAB Datasets using GCN backbone, at hidden dimensions of 16 and 32. }
    \label{figure4}
\end{figure*} 

In the experiments, we evaluated the robustness of the proposed method amidst varying channel conditions. Prior to transmission, we implemented signal power normalization. During the training phase, a consistent SNR of 5 dB was maintained, whereas for testing, the SNR was methodically varied within a range extending from -15 dB to 25 dB. Two distinct experimental sets were executed, each with a batch size of 128. The first set operated with a hidden dimension of 16, while in the second, this parameter was augmented to 32.

Fig. \ref{figure4} graphically illustrates the inference performance of the methods under evaluation across different channel quality scenarios. A discernible trend was noted, wherein the inference accuracy of all three methods demonstrated progressive enhancement in conjunction with rising SNR levels during testing, culminating in a plateau. Specifically, Fig. \ref{figure4} (a) delineates the variation of classification accuracy relative to SNR for the three methods on the PROTEINS dataset. Our proposed GIB-based method, alongside the ASAP method, showcased robust classification performance, with the former exhibiting particularly notable efficacy. The InfoGraph method achieved commendable classification accuracy, especially under near-ideal channel conditions and with a higher hidden dimension. Fig. \ref{figure4} (b) presents the experimental outcomes on the COLLAB dataset, a social network dataset characterized by complex topology and diverse node interrelations. The presence of densely connected communities within such networks significantly influences the graph's structural information, which in turn impacts the classification task. Both baseline methods exhibited suboptimal performance on this dataset, attributable to their inadequate consideration of graph topology. As explicated in Section III, the connectivity loss 
$\mathcal{L}_{con}$
  steers the model towards heightened consideration of the graph's structural characteristics, enhancing the performance of our approach.

\begin{figure*}[t]
    \centering
    \begin{subfigure}[b]{0.4\textwidth}
    \includegraphics[width=\textwidth]{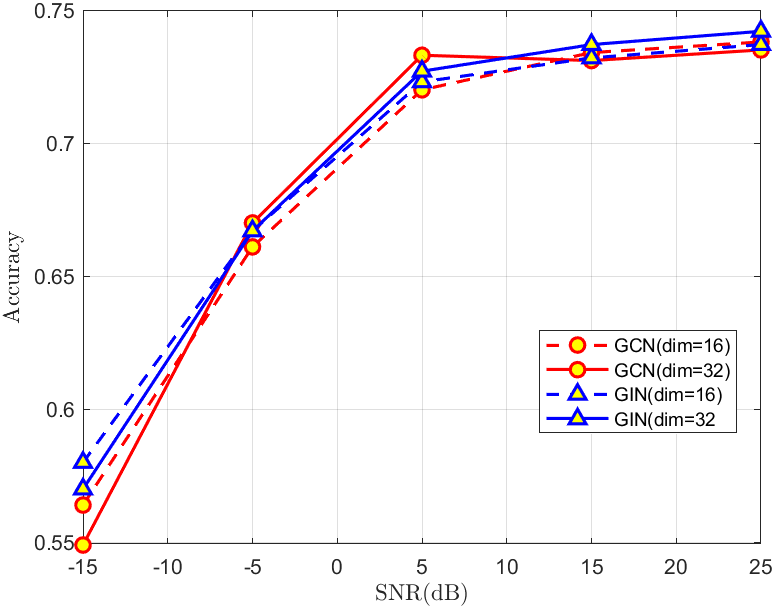}
    \captionsetup{justification=centering}
    \caption{Dataset: PROTEINS}
    \end{subfigure} 
    \hspace{0.1\textwidth}
    \begin{subfigure}[b]{0.4\textwidth}
    \includegraphics[width=\textwidth]{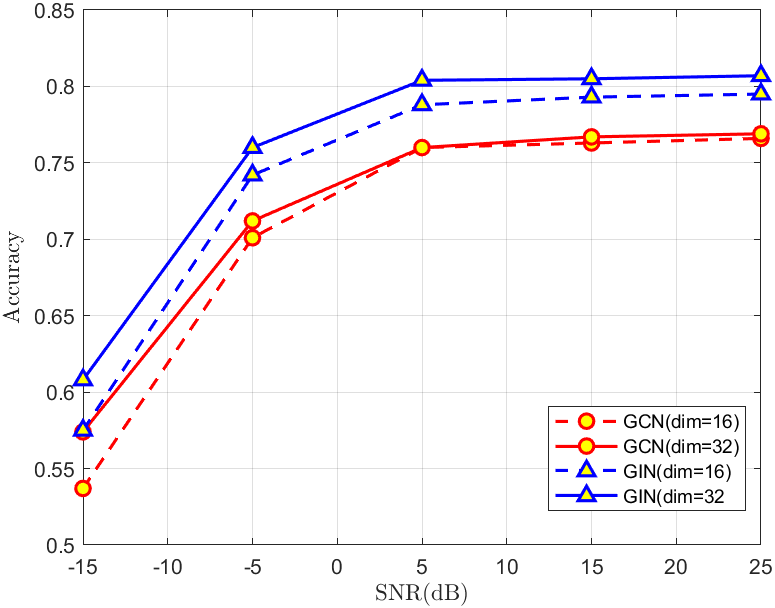}
    \captionsetup{justification=centering}
    \caption{Dataset: COLLAB}
    \end{subfigure}
    \caption{Classification accuracy variation with SNR for the proposed method on PROTEINS and COLLAB datasets, utilizing GCN and GIN backbones at hidden dimensions of 16 and 32.}
    \label{figure5}
\end{figure*} 

Fig. \ref{figure5} compares the graph classification performance between our method, which employs the GIN as the backbone, and the alternative method utilizing the GCN. These experiments were conducted on both the PROTEINS and COLLAB datasets, with hidden dimensions set to 16 and 32, respectively. Notably, in the realm of graph classification, the GIN-based method surpassed its GCN-based counterpart. The GCN, a prevalent choice for graph data processing, updates node representations through uniform aggregation of neighboring nodes. In contrast, GIN adopts a distinct methodology, where each node initially aggregates its feature with those of its neighbors via weighted summation. This non-commutative aggregation process, which is indifferent to the order of nodes, imparts to the GIN model an invariance to graph isomorphism, thus enabling it to effectively capture both the structure and global information inherent in the graph.

Given the specific focus on graph classification in this study, the adoption of GIN as the backbone is recommended. However, for a broader spectrum of tasks, the selection of GNNs with varying intrinsic characteristics as backbones can significantly bolster task-specific performance.

\subsection{Performance of GIB-enabled Task-Oriented Communication Systems Utilizing Vector Quantization}

In this part, the performance of the proposed discrete communication system, which incorporates digital codebooks, is rigorously evaluated. Our experiments are structured to assess system robustness under varying probabilities of correct transmission. Here, the probability of correct transmission, denoted as 
$\varepsilon$, signifies the likelihood that the index received by the receiver accurately corresponds with that transmitted by the sender. Consequently, higher values of 
$\varepsilon$ are indicative of superior channel quality. During the model training phase, 
$\varepsilon$ is fixed at 0.94, whereas for testing, an array of 
$\varepsilon$ values, specifically [0.90, 0.92, 0.94, 0.96, 0.98], are examined.

\begin{figure*}[t]
    \centering
    \begin{subfigure}[b]{0.4\textwidth}
    \includegraphics[width=\textwidth]{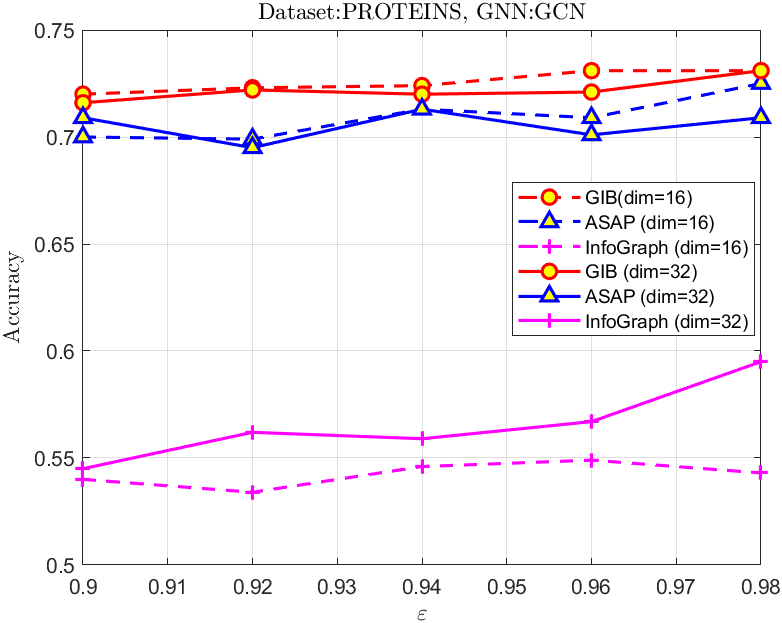}
    \captionsetup{justification=centering}
    \caption{PROTEINS}
    \end{subfigure}
    \hspace{0.1\textwidth}
    \begin{subfigure}[b]{0.4\textwidth}
    \includegraphics[width=\textwidth]{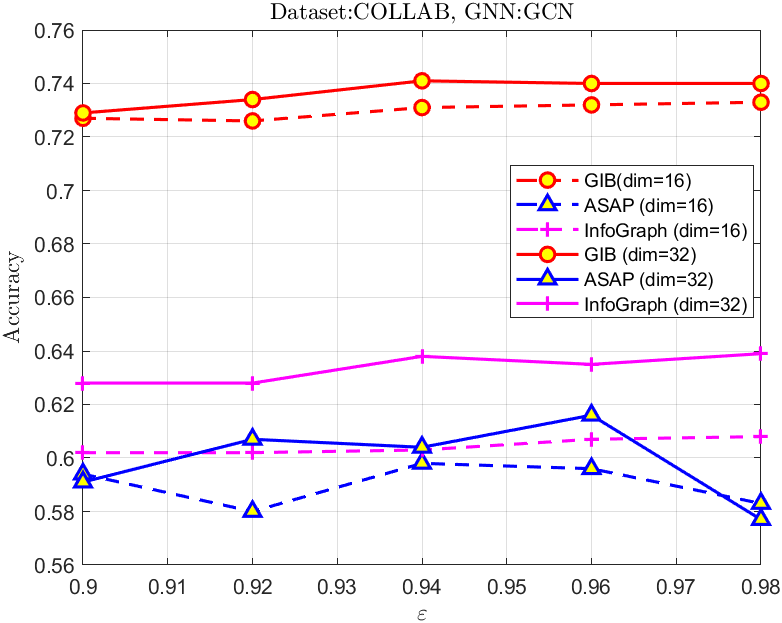}
    \captionsetup{justification=centering}
    \caption{COLLAB}
    \end{subfigure}
    \caption{Classification accuracy variation with probability of correct transmission ($\varepsilon$) for three methods on PROTEINS and COLLAB datasets using GCN backbone at hidden dimensions of 16 and 32. }
    \label{figure6}
\end{figure*}

Fig. \ref{figure6} graphically represents the classification performance of all three evaluated methods as a function of 
$\varepsilon$, with the hidden dimensions set at 16 and 32 respectively. The results delineated in these figures unequivocally demonstrate that our method significantly outperforms the baseline methods under various channel quality scenarios. The integration of InfoGraph with VQ is observed to be substantially ineffective for graph classification tasks within digital communication systems. The ASAP method exhibits moderate effectiveness on the PROTEINS dataset; however, its efficacy is markedly diminished when applied to the more complex COLLAB dataset. As illustrated in Figure \ref{figure6} (b), its classification accuracy is confined to approximately 0.6 or lower, indicating that minor transmission errors exert negligible impact on the outcomes. The curve illustrating the fluctuation of classification accuracy in relation to 
$\varepsilon$ is characterized by irregular and unpredictable oscillations. In stark contrast, our proposed method demonstrates robust performance across both datasets, with classification accuracy exhibiting a modest but consistent upward trend in line with improvements in channel quality. This outcome aligns well with our initial hypotheses and theoretical underpinnings.

\begin{figure*}[t]
    \centering
    \begin{subfigure}[b]{0.4\textwidth}
    \includegraphics[width=\textwidth]{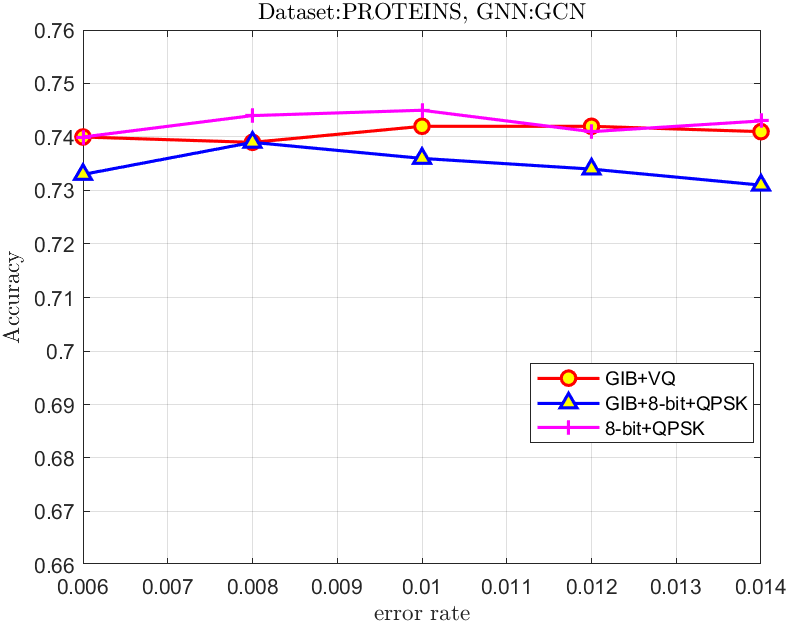}
    \captionsetup{justification=centering}
    \caption{PROTEINS}
    \end{subfigure}
    \hspace{0.1\textwidth}
    \begin{subfigure}[b]{0.4\textwidth}
    \includegraphics[width=\textwidth]{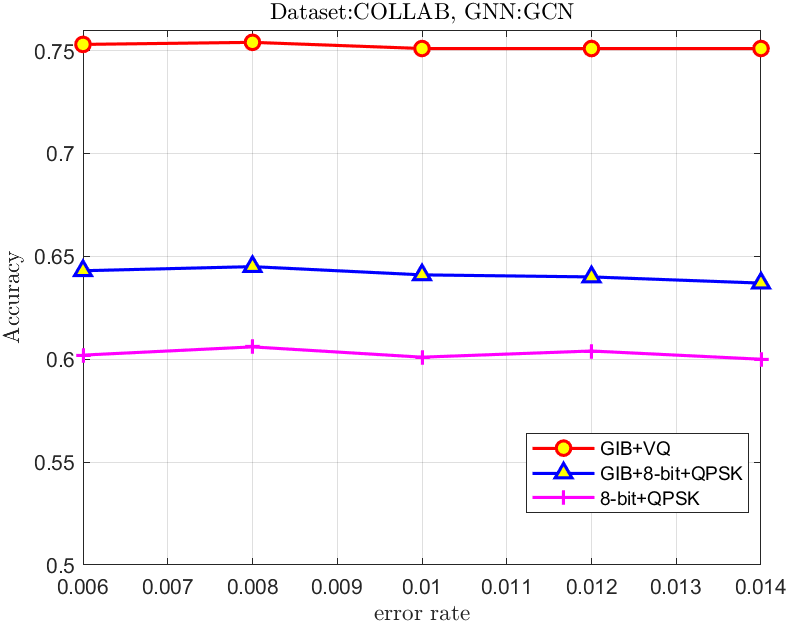}
    \captionsetup{justification=centering}
    \caption{COLLAB}
    \end{subfigure}
    \caption{Classification accuracy of three methods varying with the symbol error rate on the PROTEINS and COLLAB datasets: GIB combined with vector quantization, GIB combined with scalar quantization, and traditional digital communication using 8-bit + QPSK modulation.}
    \label{8bit-vq}
\end{figure*}

To further evaluate the performance of our proposed task-oriented digital communication system based on the GIB and VQ, we compared it against traditional digital communication method. Additionally, to validate the effectiveness of the VQ mechanism in task-oriented graph data transmission scenarios, we conducted experiment combining GIB with 8-bit scalar quantization followed by Quadrature Phase Shift Keying (QPSK) modulation.
Given the non-uniformity in the dimensionality of the signals produced by these methods, we apply the same error rate per symbol to all methods to ensure that the assessment is based on the methods' ability to handle equivalent levels of channel-induced errors, rather than on their inherent dimensions. Models were trained with a fixed error rate of 0.01 and tested across a range of error rates from 0.06 to 0.014. These experiments were carried out on two datasets, PROTEINS and COLLAB, respectively.
The experimental results are shown in Fig. \ref{8bit-vq}.

On PROTEINS, all methods exhibited comparable outcomes. This is because the PROTEINS data set is relatively simple and does not require a more elaborate symbolic representation. However, on COLLAB, the performance of the GIB based methods is significantly better than that of traditional digital communication method, which reflects the effectiveness of GIB for extracting task-related information on complex graph data. Between the two GIB-based methods, vector quantization significantly outperforms the traditional digitization method, highlighting its advantages in handling complex datasets with higher dimensionality and complexity.
The superior performance of the proposed task-oriented digital communication system based on the GIB and VQ on the COLLAB dataset highlights its potential advantages over traditional communication systems and traditional digital schemes, particularly in scenarios where data is characterized by high dimensionality and intricate patterns. 
\\

\subsection{Ablation Study}

\begin{table}[t]
\centering
\caption{Impact of different hidden dimensions on the performance of GIB and GIB without $\mathcal{L}_{MI}$ on PROTEINS dataset.}
\small
\begin{tabular}{c|c|c|c|c|c|c}
\toprule
\multicolumn{2}{c|}{SNR} & -15 dB & -5 dB & 5 dB & 15 dB & 25 dB \\
\midrule
\multirow{3}{*}{GIB} & dim-16 & 0.564 & 0.661 & 0.720 & 0.734 & 0.738 \\
                     & dim-32 & 0.549 & 0.670 & 0.733 & 0.731 & 0.735 \\
                     & \textbf{diff} & \textbf{-0.015} & \textbf{0.009} & \textbf{0.013} & \textbf{-0.003} & \textbf{-0.003} \\
\midrule
GIB & dim-16 & 0.544 & 0.634 & 0.717 & 0.728 & 0.733 \\
without & dim-32 & 0.535 & 0.640 & 0.730 & 0.731 & 0.734 \\
$\mathcal{L}_{MI}$ & \textbf{diff} & \textbf{-0.009} & \textbf{0.006} & \textbf{0.013} & \textbf{0.003} & \textbf{0.001} \\
\bottomrule
\end{tabular}
\label{PROTEINS_Lmi}
\end{table}

\begin{table}[t]
\centering
\caption{Impact of different hidden dimensions on the performance of GIB and GIB without $\mathcal{L}_{MI}$ on COLLAB dataset.}
\small
\begin{tabular}{c|c|c|c|c|c|c}
\toprule
\multicolumn{2}{c|}{SNR} & -15 dB & -5 dB & 5 dB & 15 dB & 25 dB \\
\midrule
\multirow{3}{*}{GIB} & dim-16 & 0.559 & 0.725 & 0.762 & 0.767 & 0.765 \\
                      & dim-32 & 0.586 & 0.723 & 0.770 & 0.774 & 0.778 \\
                      & \textbf{diff} & \textbf{0.027} & \textbf{-0.002} & \textbf{0.008} & \textbf{0.007} & \textbf{0.013} \\
\midrule
GIB              & dim-16 & 0.537 & 0.701 & 0.758 & 0.763 & 0.765 \\
without           & dim-32 & 0.571 & 0.720 & 0.769 & 0.773 & 0.776 \\
$\mathcal{L}_{MI}$ & \textbf{diff} & \textbf{0.034} & \textbf{0.019} & \textbf{0.011} & \textbf{0.010} & \textbf{0.011} \\
\bottomrule
\end{tabular}
\label{COLLAB_Lmi}
\end{table}

To empirically ascertain the contribution of the individual components in our proposed methodology, we executed a series of ablation studies. Specifically, we extracted $\mathcal{L}_{MI}$  and $\mathcal{L}_{con}$ from the loss function, thereby deriving two distinct variants of our method: 
GIB lacking $\mathcal{L}_{MI}$
and GIB lacking $\mathcal{L}_{con}$. 

\subsubsection{Effectiveness of $\mathcal{L}_{MI}$}

Observations from Tables \ref{PROTEINS_Lmi} and \ref{COLLAB_Lmi} indicate that, across both datasets, the system based on GIB and its variant without the $\mathcal{L}_{MI}$ component consistently exhibit superior performance at a hidden dimension of 32 compared to a hidden dimension of 16. 
This is because an increase in the hidden layer dimension increases the capacity of the model, which means that the network can capture more details in the data. Especially for graph data with complex structures, higher model capacity can help the model learn richer representations and thus improve performance. 

It is worth noting that the size of hidden dimension has a more significant effect on the performance of the variant without $\mathcal{L}_{MI}$ than GIB. 
The absence of $\mathcal{L}_{MI}$ in the variant causes the model to focus more on maximizing mutual information between the input graph data and the extracted representations, while neglecting the aspect of compression. Consequently, the extracted representations contain much redundant information that takes up space for useful information. Therefore, as the hidden dimension increases, the model capacity increases. This allows the model to capture richer information in the graph data and more information needed for the task. In other words, GIB takes into account information compression, meaning it can achieve good performance even with relatively smaller model capacity.

\begin{table}[t]
\centering
\caption{Task accuracy and standard deviation of GIB and GIB without $\mathcal{L}_{con}$ on the PROTEINS dataset at the 16-dimensional hidden dimension.}
\resizebox{0.5\textwidth}{!}{
\begin{tabular}{c|c|c|c|c|c|c}
\hline
\multicolumn{2}{c|}{SNR} & -15 dB & -5 dB & 5 dB & 15 dB & 25 dB \\ \hline
\multirow{2}{*}{GIB}& mean& 0.564 & 0.661 & 0.72 & 0.734 & 0.738 \\ 
& stdev & 0.049 & 0.050 & 0.036 & 0.028 & 0.027 \\ \hline

GIB & mean & 0.532 & 0.648 & 0.724 & 0.737 & 0.725 \\ 
without $\mathcal{L}_{con}$ & stdev & 0.052 & 0.058 & 0.039 & 0.033 & 0.030 \\ \hline
\end{tabular}
}
\label{PROTEINS_Lcon}
\end{table}

\begin{table}[t]
\centering
\caption{Task accuracy and standard deviation of GIB and GIB without $\mathcal{L}_{con}$ on the COLLAB dataset at the 16-dimensional hidden dimension.}
\resizebox{0.5\textwidth}{!}{
\begin{tabular}{c|c|c|c|c|c|c}
\hline
\multicolumn{2}{c|}{SNR} & -15 dB & -5 dB & 5 dB & 15 dB & 25 dB \\ \hline
\multirow{2}{*}{GIB}& mean& 0.537 &	0.701 &	0.760 &	0.763 &	0.768 
 \\ 
& stdev & 0.022 & 0.012 & 0.012 & 0.014 & 0.013 \\ \hline

GIB & mean & 0.541 & 0.699 & 0.752 & 0.760 & 0.763 \\ 
without $\mathcal{L}_{con}$ & stdev & 0.018 & 0.024 & 0.016 & 0.016 & 0.013 \\ \hline
\end{tabular}
}
\label{COLLAB_Lcon}
\end{table}

\subsubsection{Effectiveness of $\mathcal{L}_{con}$}
Furthermore, the absence of the connectivity loss constraint, 
$\mathcal{L}_{con}$
 , not only detrimentally impacts system performance on certain datasets but also instigates volatility in node assignments, subsequently leading to inconsistent graph classification results. 
To illustrate the significance of connectivity loss and its impact on system reliability, we adopt the arithmetic mean of outcomes derived from a 10-fold cross-validation process as the benchmark for task execution accuracy. Complementarily, the standard deviation computed from these same 10 iterations serves as the metric to gauge the consistency and robustness of our system's performance under different communication conditions. 
It becomes evident upon analysis that the standard deviation associated with GIB added $\mathcal{L}_{con}$ exhibits a reduction in comparison to its variant without $\mathcal{L}_{con}$ under various channel conditions. This result suggests that the connectivity loss we incorporate plays a critical role in improving the robustness and stability of the system performance.

\begin{table}[t]
\centering
\caption{Influence of Mutual Information Weight $\beta$ on Graph Classification Accuracy}
\small
\begin{tabular}{c|c|c|c|c|c}
\hline
\diagbox{\quad $\beta$}{SNR} & -15 dB & -5 dB & 5 dB & 15 dB & 25 dB \\ \hline
0.01 & 0.530 & 0.648 & 0.706 & 0.722 & 0.731 \\ \hline
0.1 & 0.564 & \textbf{0.661} & 0.720 & \textbf{0.734} & \textbf{0.738} \\ \hline
0.3 & \textbf{0.575} & 0.650 & \textbf{0.727} & 0.730 & 0.733 \\ \hline
0.5 & 0.552 & 0.660 & 0.718 & 0.724 & 0.727 \\ \hline
0.7 & 0.571 & 0.643 & 0.722 & 0.728 & 0.736 \\ \hline
\end{tabular}
\label{tab5}
\end{table}

\subsubsection{Influence of Variations in 
$\beta$ on System Performance}
The hyperparameter $\beta$, associated with the mutual information term, plays a pivotal role in modulating the extent of mutual information between the graph features and the subgraph features, thereby reflecting the relative importance of the graph's global structure versus its local features. To explore this interplay, we conducted experiments assessing the impact of varying $\beta$ values on classification accuracy, with the findings presented in Table \ref{tab5}. The experimental data reveal that the system attains optimal performance at a moderate MI factor level, in contrast to either extremely high or low levels.

When $\beta$ is set to a low value, the model predominantly accentuates local features, while marginalizing the global structural information of the graph. This can lead to the model's failure in capturing critical global relationships within the graph, culminating in a diminished classification accuracy. Inversely, a high $\beta$ value results in the model prioritizing the global graph structure at the expense of local features. Such an overemphasis on the macro structure of the graph renders the classifier less receptive to nuanced local features, thereby constraining the model's task execution capabilities. By selecting a moderate value for $\beta$, the model achieves an equilibrium between harnessing the global structure and incorporating local features. This equilibrium enables the model to effectively utilize the intrinsic global structure of the graph while concurrently considering the node-specific local features. The resultant synergistic integration of global and local features thus significantly enhances the model's performance in graph classification tasks.

\section{Discussion}
Task-oriented graph data transmission is a promising research avenue, especially relevant in smart city scenarios such as traffic management and environmental monitoring, where distributed edge computing plays a crucial role.  In these contexts, numerous sensors and devices across the city communicate with central servers for data processing.  This communication technique significantly reduces bandwidth usage while maintaining accurate inference capabilities, making it highly suitable for real-time applications in urban settings.  To effectively translate the theoretical foundations of the GIB approach into tangible applications, it is essential to demonstrate its practical utility in addressing real-world, task-oriented communication challenges. Here are some heuristics worth exploring.

\subsection{Heuristics for Inference Goal Formulation} Real-world scenarios encompass a broad spectrum of inference objectives, contingent upon the specific application context. For example, in the realm of autonomous vehicles, the focus might be on detecting and responding to traffic signals and obstacles. In contrast, industrial IoT may concentrate on predicting equipment failures. The GIB framework is designed to fine-tune the communication process, thereby amplifying the precision and efficiency tailored to these particular tasks. The GIB model can be crafted to cater to either uni-task or multi-tasks. While task-specific models are optimized for a unique objective, task-agnostic models can be expanded to address a variety of tasks through strategic architectural enhancements and transfer learning techniques. This approach necessitates training the model across a diverse array of tasks and datasets, fostering its ability to generalize across varied communication scenarios.

\subsection{Heuristics for GIB Model Training} The GIB model should be trained on data that mirrors the complexities of its target real-world task. This process entails amassing comprehensive datasets that encompass both raw input data, such as sensor readings or visual imagery, and the relevant task-specific labels, such as classifications of objects or indicators of faults. It is imperative that these data are meticulously annotated and preprocessed to construct a graph that encapsulates the interconnections and interdependencies among various data elements. The efficacy of GIB framework should be substantiated through rigorous experimentation with datasets that pertain to different task-oriented communication applications. For instance,  industrial datasets could be utilized to train the model in predictive maintenance tasks, employing sensor data to anticipate equipment failures. Additionally, deploying the GIB framework on medical imaging datasets could enhance communication efficiency in tasks such as diagnosing diseases and monitoring patient conditions. 

\subsection{Heuristics for Training Dataset Construction}
To construct a training dataset for tasks without a standard dataset, follow these steps: First, clearly specify the task, such as node classification or link prediction; second, collect raw data needed for the graph-based task, typically in the form of nodes, edges, and node features; third, preprocess the data suitable for graph-based learning by normalizing node features, removing self-loops, and converting the graph to an appropriate format (e.g., adjacency matrix, edge list); Construct graph representations to serve as inputs to GNN, typically involving adjacency matrices and node feature matrices; Generate pairs of input graphs and target outputs specific to the task. For instance, for node classification, the input is subgraphs centered around each node with the output being the class label of the central node for node classification tasks, while for link prediction tasks, the input is subgraphs around a candidate link with the target being a binary label indicating whether the link exists; Implement GIB to identify and retain the most informative parts of the graph while discarding unnecessary information by encoding the graph data into a compressed representation and decoding it to approximate the original or task-specific output; Finally,  divide the dataset to ensure the model generalizes well to unseen data. Additionally, feature engineering, such as aggregating neighborhood information and generating higher-order features, can enrich node features. Data augmentation can also be used to improve the model's robustness in data-hungry cases.

\section{Conclusion}

In this study, we explored GIB-based task-oriented communication for graph data transmission, focusing on optimizing mutual information between received codewords and the task goal while reducing the mutual information with the original graph representation. Utilizing variational approximation, Monte Carlo sampling, and MINE, we devised a workable objective function, addressing the challenge of computing mutual information for irregular graph data. We also incorporated a connectivity loss term to account for community structure in graph data, which improved subgraph selection and stabilized training, as confirmed by our experiments. Furthermore, we adapted our method for digital transmission using vector quantization. Our tests across various SDC channel qualities consistently showed strong performance, underscoring the method's adaptability and efficacy. Future work will explore more practical application areas and include hands-on experiments with real-world data.

\bibliographystyle{IEEEtran} 
\bibliography{Task-Oriented_Communication_for_Graph_Data}

\begin{thebibliography}{10}
\providecommand{\url}[1]{#1}
\csname url@samestyle\endcsname
\providecommand{\newblock}{\relax}
\providecommand{\bibinfo}[2]{#2}
\providecommand{\BIBentrySTDinterwordspacing}{\spaceskip=0pt\relax}
\providecommand{\BIBentryALTinterwordstretchfactor}{4}
\providecommand{\BIBentryALTinterwordspacing}{\spaceskip=\fontdimen2\font plus
\BIBentryALTinterwordstretchfactor\fontdimen3\font minus \fontdimen4\font\relax}
\providecommand{\BIBforeignlanguage}[2]{{%
\expandafter\ifx\csname l@#1\endcsname\relax
\typeout{** WARNING: IEEEtran.bst: No hyphenation pattern has been}%
\typeout{** loaded for the language `#1'. Using the pattern for}%
\typeout{** the default language instead.}%
\else
\language=\csname l@#1\endcsname
\fi
#2}}
\providecommand{\BIBdecl}{\relax}
\BIBdecl

\bibitem{ref002}
M.~Mohsenivatani, S.~Ali, V.~Ranasinghe, N.~Rajatheva, and M.~Latva-Aho, ``Graph representation learning for wireless communications,'' \emph{IEEE Commun. Mag.}, vol.~62, no.~1, pp. 141--147, 2024.

\bibitem{9027556}
Y.~Tian, Y.~Niu, J.~Yan, and F.~Tian, ``Inferring private attributes based on graph convolutional neural network in social networks,'' in \emph{Proc. Int. Conf. Netw. Netw. Appl. (NaNA)}, Daegu, South Korea, Mar. 2019, pp. 186--190.

\bibitem{7976923}
S.~Shaikh, S.~Rathi, and P.~Janrao, ``Recommendation system in e-commerce websites: A graph based approached,'' in \emph{Proc. IEEE 7th Int. Advance Comput. Conf. (IACC)}, Daegu, Korea, Jul. 2017, pp. 931--934.

\bibitem{8047276}
Q.~Wang, Z.~Mao, B.~Wang, and L.~Guo, ``Knowledge graph embedding: A survey of approaches and applications,'' \emph{{IEEE} Trans. Knowl. Data Eng.}, vol.~29, no.~12, pp. 2724--2743, Sep. 2017.

\bibitem{7965747}
Y.~Duan, L.~Shao, G.~Hu, Z.~Zhou, Q.~Zou, and Z.~Lin, ``Specifying architecture of knowledge graph with data graph, information graph, knowledge graph and wisdom graph,'' in \emph{Proc. IEEE 15th Int. Conf. Softw. Eng. Res. Manage. Appl. (SERA)}, London, United Kingdom, Jul. 2017, pp. 327--332.

\bibitem{4684911}
W.~W.~M. Lam and K.~C.~C. Chan, ``A graph mining algorithm for classifying chemical compounds,'' in \emph{Proc. IEEE Int. Conf. Bioinf. Biomedicine}, Philadephia, Pennsylvania, USA, Nov. 2008, pp. 321--324.

\bibitem{ref001}
M.~Henaff, J.~Bruna, and Y.~LeCun, ``Deep convolutional networks on graph-structured data,'' \emph{ArXiv: 1506.05163}, Jun. 2015.

\bibitem{Gndz2022BeyondTB}
D.~Gündüz, Z.~Qin, I.~E. Aguerri, H.~S. Dhillon, Z.~Yang, A.~Yener, K.~K. Wong, and C.-B. Chae, ``Beyond transmitting bits: Context, semantics, and task-oriented communications,'' \emph{{IEEE} J. Sel. Areas Commun.}, vol.~41, pp. 5--41, Nov. 2022.

\bibitem{Xie2020DeepLE}
H.~Xie, Z.~Qin, G.~Y. Li, and B.-H. Juang, ``Deep learning enabled semantic communication systems,'' \emph{{IEEE} Trans. Signal Process.}, vol.~69, pp. 2663--2675, Apr. 2021.

\bibitem{Luo2022SemanticCO}
X.~Luo, H.-H. Chen, and Q.~Guo, ``Semantic communications: Overview, open issues, and future research directions,'' \emph{{IEEE} Wireless Commun.}, vol.~29, pp. 210--219, Jan. 2022.

\bibitem{Shao2021LearningTC}
J.~Shao, Y.~Mao, and J.~Zhang, ``Learning task-oriented communication for edge inference: An information bottleneck approach,'' \emph{{IEEE} J. Sel. Areas Commun.}, vol.~40, pp. 197--211, Nov. 2021.

\bibitem{Weng2021SemanticCF}
Z.~Weng, Z.~Qin, and G.~Y. Li, ``Semantic communications for speech recognition,'' in \emph{Proc. IEEE Global Commun. Conf. (GLOBECOM)}, Feb. 2022, pp. 1--6.

\bibitem{Shao2021TaskOrientedCF}
J.~Shao, Y.~Mao, and J.~Zhang, ``Task-oriented communication for multidevice cooperative edge inference,'' \emph{{IEEE} Trans. Wireless Commun.}, vol.~22, pp. 73--87, Jul. 2022.

\bibitem{Farsad2018DeepLF}
N.~Farsad, M.~Rao, and A.~J. Goldsmith, ``Deep learning for joint source-channel coding of text,'' in \emph{Proc. IEEE Int. Conf. Acoust. Speech Signal Process. (ICASSP)}, Calgary, AB, Canada, Sep. 2018, pp. 2326--2330.

\bibitem{10177738}
S.~Guo, Y.~Wang, S.~Li, and N.~Saeed, ``Semantic importance-aware communications using pre-trained language models,'' \emph{{IEEE} Commun. Lett.}, vol.~27, no.~9, pp. 2328--2332, Jul. 2023.

\bibitem{Bourtsoulatze2018DeepJS}
E.~Bourtsoulatze, D.~B. Kurka, and D.~G{\"u}nd{\"u}z, ``Deep joint source-channel coding for wireless image transmission,'' \emph{{IEEE} Trans. on Cogn. Commun. Netw.}, vol.~5, pp. 567--579, Apr. 2018.

\bibitem{Kurka2019DeepJSCCfDJ}
D.~B. Kurka and D.~G{\"u}nd{\"u}z, ``Deep{JSCC}-f: Deep joint source-channel coding of images with feedback,'' \emph{{IEEE} J. Sel. Areas Inf. Theory}, vol.~1, pp. 178--193, Apr. 2020.

\bibitem{Kang2021TaskOrientedIT}
X.~Kang, B.~Song, J.~Guo, Z.~Qin, and F.~R. Yu, ``Task-oriented image transmission for scene classification in unmanned aerial systems,'' \emph{{IEEE} Trans. Commun.}, vol.~70, pp. 5181--5192, Jun. 2022.

\bibitem{Li2022GraphSC}
P.~Li, N.~Shlezinger, H.~Zhang, B.~Wang, and Y.~C. Eldar, ``Graph signal compression by joint quantization and sampling,'' \emph{{IEEE} Trans. Signal Process.}, vol.~70, pp. 4512--4527, Sep. 2022.

\bibitem{9416834}
F.~Xia, K.~Sun, S.~Yu, A.~Aziz, L.~Wan, S.~Pan, and H.~Liu, ``Graph learning: A survey,'' \emph{{IEEE} Trans. Artif. Intell.}, vol.~2, no.~2, pp. 109--127, Apr. 2021.

\bibitem{Wang2023PrivacyPreservingTS}
Y.~Wang, S.~Guo, Y.~Deng, H.~Zhang, and Y.~Fang, ``Privacy-preserving task-oriented semantic communications against model inversion attacks,'' \emph{IEEE Trans. Wireless Commun.}, pp. 1--1, Mar. 2024.

\bibitem{DBLP:journals/corr/physics-0004057}
N.~Tishby, F.~C.~N. Pereira, and W.~Bialek, ``The information bottleneck method,'' in \emph{Proc. Annu. Allerton Conf. Commun. Control Comput.}, Oct. 1999, p. 368–377.

\bibitem{Tishby2015DeepLA}
N.~Tishby and N.~Zaslavsky, ``Deep learning and the information bottleneck principle,'' in \emph{Proc. IEEE Inf. Theory Workshop (ITW)}, Jerusalem, Israel, Jun. 2015, pp. 1--5.

\bibitem{Wu2022HandlingDS}
Q.~Wu, H.~Zhang, J.~Yan, and D.~P. Wipf, ``Handling distribution shifts on graphs: An invariance perspective,'' in \emph{Proc. Int. Conf. Learn. Represent., {ICLR}}, Virtual, Feb. 2022.

\bibitem{MACFL2022TWC}
C.~Feng, H.~H. Yang, D.~Hu, Z.~Zhao, T.~Q.~S. Quek, and G.~Min, ``Mobility-aware cluster federated learning in hierarchical wireless networks,'' \emph{IEEE Transactions on Wireless Communications}, vol.~21, no.~10, pp. 8441--8458, Oct. 2022.

\bibitem{DBLP:conf/nips/WuRLL20}
T.~Wu, H.~Ren, P.~Li, and J.~Leskovec, ``Graph information bottleneck,'' in \emph{Advances Neural Inf. Process. Syst. (NeurIPS)}, vol.~33, Virtual, 2020, pp. 437--448.

\bibitem{Wu2019ACS}
Z.~Wu, S.~Pan, F.~Chen, G.~Long, C.~Zhang, and P.~S. Yu, ``A comprehensive survey on graph neural networks,'' \emph{{IEEE} Trans. Neural Netw. Learn. Syst.}, vol.~32, pp. 4--24, Mar. 2020.

\bibitem{LargeGNN}
G.~Zhao, Q.~Wang, F.~Yao, Y.~Zhang, and G.~Yu, ``Survey on large-scale graph neural network systems,'' \emph{J. Software}, vol.~33, no.~1, pp. 150--170, 2022.

\bibitem{DBLP:conf/iclr/AlemiFD017}
A.~A. Alemi, I.~Fischer, J.~V. Dillon, and K.~Murphy, ``Deep variational information bottleneck,'' in \emph{Proc. Int. Conf. Learn. Represent., {ICLR}}, Apr. 2017, pp. 1--19.

\bibitem{Belghazi2018MutualIN}
M.~I. Belghazi, A.~Baratin, S.~Rajeswar, S.~Ozair, Y.~Bengio, R.~D. Hjelm, and A.~C. Courville, ``Mutual information neural estimation,'' in \emph{Proc. Int. Conf. on Mach. Learn. (ICML)}, vol.~80, Stockholmsm{\"{a}}ssan, Stockholm, Sweden, Jul. 2018, pp. 530--539.

\bibitem{yu2021graph}
J.~Yu, T.~Xu, Y.~Rong, Y.~Bian, J.~Huang, and R.~He, ``Graph information bottleneck for subgraph recognition,'' in \emph{Proc. Int. Conf. Learn. Represent., {ICLR}}, Oct. 2020.

\bibitem{Xie2022RobustIB}
S.~Xie, S.~Ma, M.~Ding, Y.~Shi, M.-F. Tang, and Y.~Wu, ``Robust information bottleneck for task-oriented communication with digital modulation,'' \emph{{IEEE} J. Sel. Areas Commun.}, vol.~41, pp. 2577--2591, Jun. 2023.

\bibitem{Bo2022LearningBJ}
Y.~Bo, Y.~Duan, S.~Shao, and M.~Tao, ``Learning based joint coding-modulation for digital semantic communication systems,'' \emph{Proc. 14th Int. Conf. Wireless Commun. Signal Process. (WCSP)}, pp. 1--6, Feb. 2023.

\bibitem{Hu2022RobustSC}
Q.~Hu, G.~Zhang, Z.~Qin, Y.~Cai, G.~Yu, and G.~Y. Li, ``Robust semantic communications with masked {VQ-VAE} enabled codebook,'' \emph{{IEEE} Trans. Wireless Commun.}, vol.~22, no.~12, pp. 8707--8722, Apr. 2023.

\bibitem{DBLP:conf/iclr/SunHV020}
F.~Sun, J.~Hoffmann, V.~Verma, and J.~Tang, ``Infograph: Unsupervised and semi-supervised graph-level representation learning via mutual information maximization,'' in \emph{Proc. Int. Conf. Learn. Represent., {ICLR}}, Addis Ababa, Ethiopia, Apr. 2020.

\bibitem{DBLP:conf/iclr/XuHLJ19}
K.~Xu, W.~Hu, J.~Leskovec, and S.~Jegelka, ``How powerful are graph neural networks?'' in \emph{Proc. Int. Conf. Learn. Represent., {ICLR}}, New Orleans, LA, USA, May 2019.

\bibitem{Ranjan2019ASAPAS}
E.~Ranjan, S.~Sanyal, and P.~P. Talukdar, ``{ASAP}: Adaptive structure aware pooling for learning hierarchical graph representations,'' in \emph{Proc. AAAI Conf. Artif. Intell.}, vol.~34, no.~4, New York, NY, USA, Feb. 2019, pp. 5470--5477.

\end{thebibliography}
\end{document}